\documentclass[10pt,twocolumn,letterpaper]{article}

\usepackage{graphbox}
\usepackage{graphicx}
\usepackage{subcaption}
\usepackage{float}
\usepackage{lscape}                                         % Useful for wide tables or figures.
\usepackage{wrapfig}

\usepackage[lined,ruled,linesnumbered]{algorithm2e}
\usepackage{animate} %% convert frames to video

\usepackage{booktabs}                   % Publication quality tables
\usepackage{multirow}

\usepackage{makecell}

\usepackage{paralist}
\usepackage{enumitem}

\usepackage{bm}                          % Make bold, italic math symbols
\usepackage{epsfig}                      % for figures
\usepackage{graphicx}                  % another package that works for figures
\usepackage{mathtools}
\usepackage{amssymb,amsmath}   % Short math guide for LaTeX ftp://ftp.ams.org/pub/tex/doc/amsmath/short-math-guide.pdf

\usepackage{units}
\usepackage{color}

\usepackage{comment}

\usepackage{url}  % Hyphenation of URLs.
\usepackage[pagebackref=true,breaklinks=true,colorlinks,bookmarks=false,linkcolor=blue,citecolor=blue]{hyperref}

\usepackage{xspace}
\usepackage[table]{xcolor}
\usepackage{setspace}

\def\etal{et~al.}			  % and others, and co-workers
\def\eg{e.g.,~}               % for example
\def\ie{i.e.,~}               % that is, in other words
\newlength\paramargin
\newlength\figmargin

\newlength\secmargin
\newlength\figcapmargin
\newlength\tabcapmargin

\newcommand{\secref}[1]{Section~\ref{sec:#1}}
\newcommand{\figref}[1]{Figure~\ref{fig:#1}} 
\newcommand{\tabref}[1]{Table~\ref{tab:#1}}
\newcommand{\eqnref}[1]{\eqref{eq:#1}}

\newcommand{\Paragraph}[1]
{\vspace{1mm} \noindent \textbf{#1}}

\long\def\ignorethis#1{}

\newcommand {\first}[1]{{\color{red}\textbf{#1}}}
\newcommand {\second}[1]{{\color{blue}\underline{#1}}}

{}

\newbox\jsavebox%

\def\xi{\mathbf{x}_i}

\graphicspath{{figure}, {example}}

\usepackage{iccv}
\usepackage{times}
\usepackage{epsfig}
\usepackage{graphicx}
\usepackage{amsmath}
\usepackage{amssymb}

\iccvfinalcopy % *** Uncomment this line for the final submission

\def\iccvPaperID{1906} % *** Enter the ICCV Paper ID here

\ificcvfinal\fi

\begin{document}

%%%%%%%%% TITLE
\title{Learning to Stylize Novel Views}

\author{Hsin-Ping Huang$^1$, Hung-Yu Tseng$^1$, Saurabh Saini$^2$, Maneesh Singh$^2$, Ming-Hsuan Yang$^{1,3,4}$\vspace{1mm}\\
$^1$UC Merced~~~$^2$Verisk Analytics~~~$^3$Google Research~~~$^4$Yonsei University\vspace{1mm} \\
\footnotesize\url{https://hhsinping.github.io/3d_scene_stylization}
\vspace{-3mm}
}

\ificcvfinal\thispagestyle{empty}\fi

\twocolumn[{
\renewcommand\twocolumn[1][]{#1}
\maketitle
\begin{center}
    \centering
    \vspace{-3mm}
    \captionsetup{type=figure}
    \includegraphics[width=\linewidth]{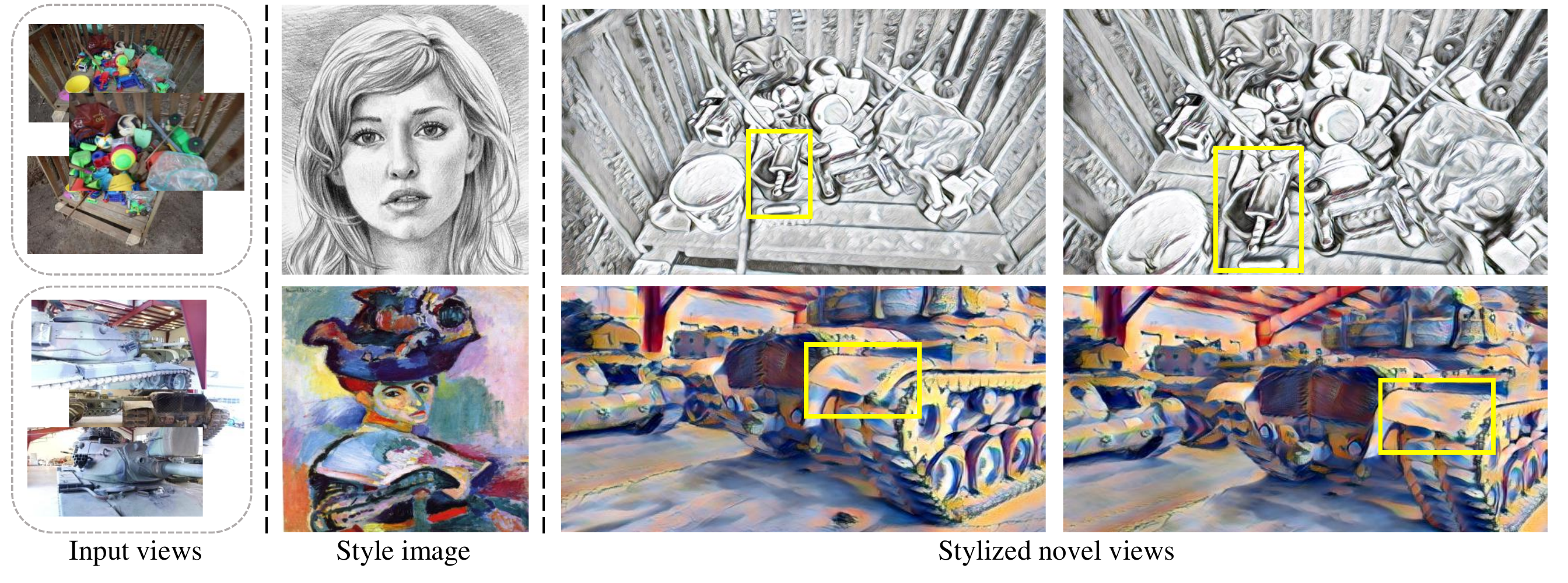}
    \vspace{-4mm}
    \captionof{figure}
    {\textbf{3D scene stylization}. Given a set of images of a 3D scene (\textit{left}) as well as a reference image of the desired style (\textit{middle}), our method is able to modify the style of the 3D scene, and synthesize images of arbitrary novel views (\textit{right}). 
    The novel view synthesis results 1) contain the desired style and 2) are consistent across various novel views, \eg the texture in the yellow boxes.}
    \label{fig:teaser}
\end{center}
}]


%%%%%%%%% ABSTRACT
\begin{abstract}
We tackle a 3D scene stylization problem --- generating stylized images of a scene from arbitrary novel views given a set of images of the same scene and a reference image of the desired style as inputs.
Direct solution of combining novel view synthesis and stylization approaches lead to results that are blurry or not consistent across different views. 
We propose a point cloud-based method for consistent 3D scene stylization. 
First, we construct the point cloud by back-projecting the image features to the 3D space.
Second, we develop point cloud aggregation modules to gather the style information of the 3D scene, and then modulate the features in the point cloud with a linear transformation matrix.
Finally, we project the transformed features to 2D space to obtain the novel views.
Experimental results on two diverse datasets of real-world scenes validate that our method generates consistent stylized novel view synthesis results against other alternative approaches. 
\end{abstract}
\vspace{-5mm}
\section{Introduction}
\label{sec:intro}

\begin{figure*}[t!]
\centering
\includegraphics[width=\linewidth]{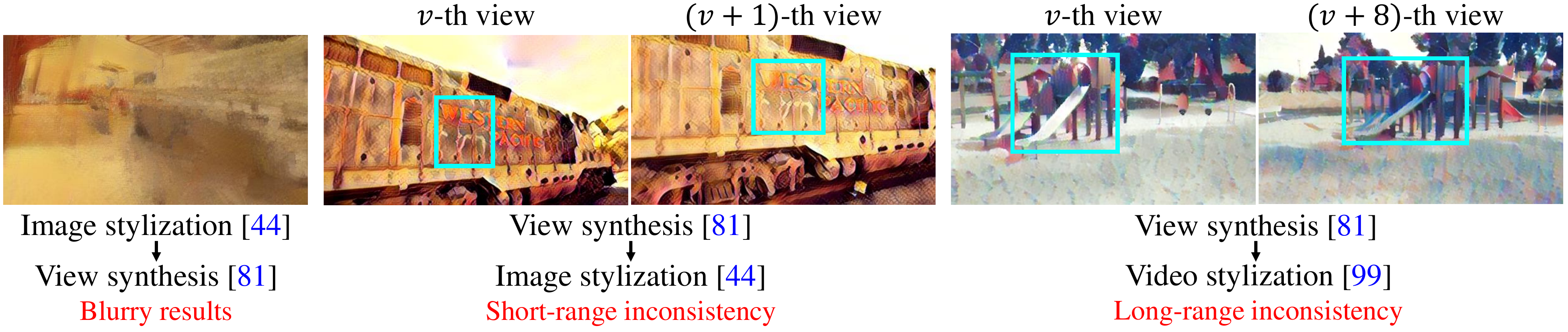}
\caption{\textbf{Motivation.} While the existing methods can be used for the 3D scene stylization task, these methods either produce blurry (image stylization $\rightarrow$ novel view synthesis), short-range inconsistent (novel view synthesis $\rightarrow$ image stylization), or long-range inconsistent (novel view synthesis $\rightarrow$ video stylization) results.}
\vspace{-2mm}
\label{fig:motivation}
\end{figure*}

Visual content creation in 3D space has recently attracted increasing attention.
Driven by the success of 3D scene representation approaches~\cite{nerf,Riegler2020SVS,kaizhang2020}, recent methods make significant progress on various content creation tasks for 3D scenes, such as semantic view synthesis~\cite{habtegebrial2020generative,huang2020semantic} and scene extrapolation~\cite{liu2020infinite}.
%
%\textcolor{red}{Most existing work focuses on reconstructing 3D scenes and produce photo-realistic novel views. The problem of manipulating/editing reconstructed 3D scenes has not been well-explored.}
In this work, we focus on the \emph{3D scene stylization} problem.
As shown in \figref{teaser}, given a set of images of a target scene and a reference image of the desired style, our goal is to render stylized images of the scene from arbitrary novel views.
3D scene stylization enables a variety of interesting virtual reality~(VR) and augmented reality~(AR) applications, \eg augment the street scene at user locations to the \textit{Cafe Terrace at Night} style by van Gogh.

Learning to modify the style of an existing 3D scene is challenging for two reasons. 
First, the synthesized novel views (\ie 2D images) of the stylized 3D scene must contain the desired style provided by the reference image.
Second, since our goal is to stylize the \emph{holistic} 3D scene, the generated novel views need to be consistent across different viewpoints for the same scene, such as the texture in the yellow boxes shown in \figref{teaser}.

To handle these challenges, one plausible solution is to combine existing novel view synthesis~\cite{Riegler2020SVS,kaizhang2020} and image stylization approaches~\cite{li2018learning,Svoboda_2020_CVPR}.
However, such straightforward approaches lead to problematic results since image stylization schemes are not designed to consider the consistency issue across different views for the same scene.
We present the examples in \figref{motivation} where
the results may be blurry if the input images of the target scene are stylized before conducting novel view synthesis.
On the other hand, if we apply image stylization after novel view synthesis, the results are not consistent across different views.
Another possible solution is to treat a series of novel view synthesis results as a \emph{video}, and use the video stylization frameworks~\cite{deng2020arbitrary,gao2020,Compound2020} to obtain temporally consistent results.
However, as shown in~\figref{motivation}, these approaches are not able to enforce long-range consistency (\ie between two far-away views) as the video stylization schemes only guarantee the short-term consistency.

In this paper, we propose a point cloud-based method for consistent 3D scene stylization.
To synthesize novel views that 1) match \emph{arbitrary} style images
%style image 
and 2) render images with consistent appearance across different views, the core idea is to operate on the 3D scene representation, \ie point cloud, of the target scene.
Given a set of input images of the target scene, we first construct the point cloud by back-projecting the image features to the 3D space according to the pre-computed 3D proxy geometry.
To transfer the style of the holistic 3D scene, we develop a point cloud transformation module.
Specifically, we use a series of point cloud aggregation modules to gather the style information of the 3D scene.
We then modulate the features in the point cloud with a linear transformation matrix~\cite{li2018learning} computed according to the style information of the point cloud and reference image.
Finally, we project the transformed features from the point cloud to the 2D space to obtain the novel view synthesis results.
Since our method synthesizes novel view images from the same stylized point cloud, the rendered results not only demonstrate the desired style, but also are consistent across different viewpoints.
%
%\textcolor{red}{With a single trained model, the proposed method can transfer \emph{arbitrary styles} to \emph{arbitrary 3D scenes} without fine-tuning, and the stylized novel views are rendered in near-real-time ($17$ fps).}

We evaluate the proposed 3D scene stylization method through extensive qualitative and quantitative studies.
The experiments are conducted on two diverse datasets of real-world scenes: Tanks and Temples~\cite{Knapitsch2017} and FVS~\cite{Riegler2020FVS}.
We conduct a user preference study to evaluate the stylization quality, \ie whether the novel view synthesis results match the style of the reference image.
In addition, we use the Learned Perceptual Image Patch Similarity~(LPIPS)~\cite{lpips} metric to measure the consistency of the results synthesized across different novel views.

We make the following contributions in this paper:
\begin{compactitem}
\item We propose a point cloud-based framework for the 3D scene stylization task.
\item We design a point cloud transformation module that learns to transfer the style from an arbitrary 2D reference image to the point cloud of a 3D scene.
\item We validate that our method produces high-quality and consistent stylized novel view synthesis results on the Tanks and Temples as well as FVS datasets.
\end{compactitem}

\section{Related Work}
\label{sec:related}
\begin{figure*}[t!]
\centering
\includegraphics[width=0.95\linewidth]{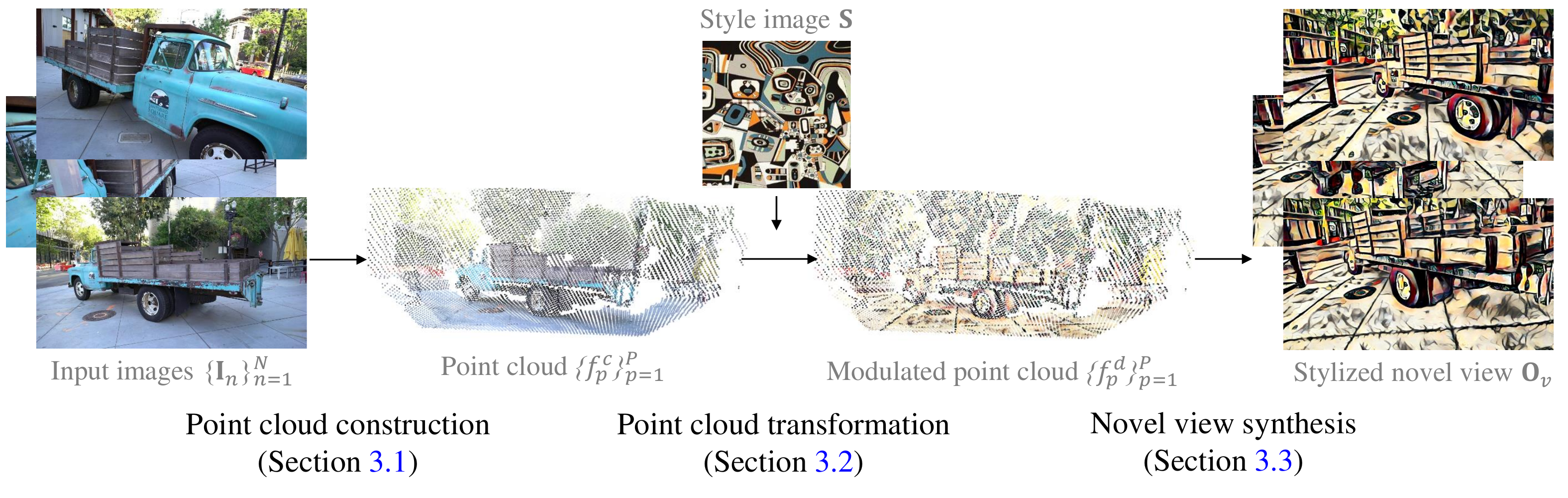}
\vspace{-1.5mm}
\caption{
\textbf{Algorithmic overview.}
The proposed method consists of three steps: 1) constructing the 3D point cloud from the set of input images $\{\mathbf{I}_n\}^N_{n=1}$, 2) transforming the point cloud according to the reference image $\mathbf{S}$ with the desired style, and 3) synthesizing the stylized image $\mathbf{O}_v$ at arbitrary novel view $v$.
The coloring of the point clouds is for visualization purposes only.
In our approach, the point clouds store the features rather than RGB values.
}
\vspace{-3mm}
\label{fig:overview}
\end{figure*}
\begin{figure}[t!]
\centering
\includegraphics[width=\linewidth]{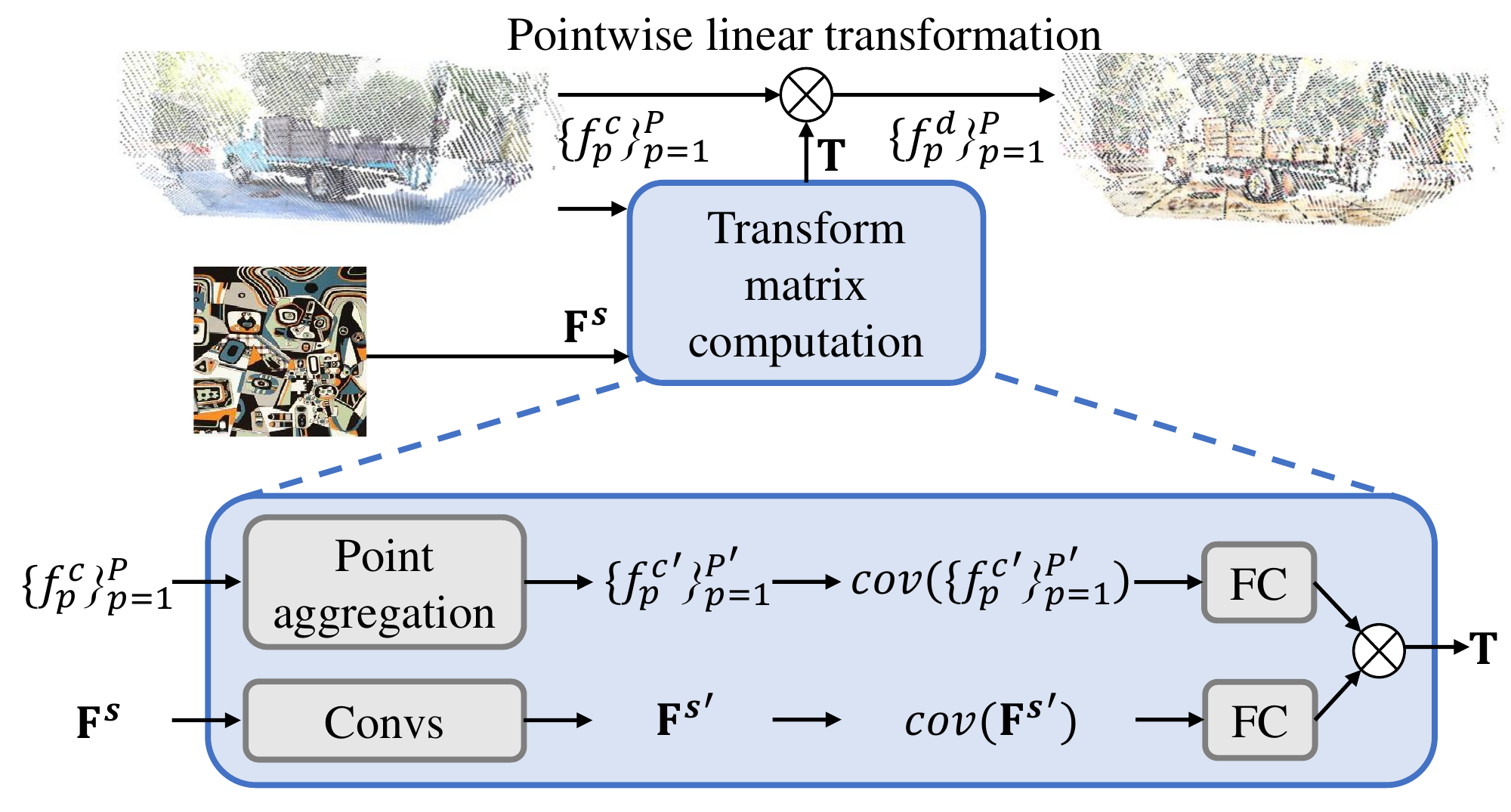}
\caption{
\textbf{Point cloud transformation.}
We model the 3D scene stylization process as the linear transformation between the constructed and stylized point clouds. 
Specifically, the constructed point cloud is modulated using the predicted linear transformation matrix $\mathbf{T}$, as described in \eqnref{transformation}.
We use a series of point cloud aggregation modules to gather the point cloud information, and the convolution layers to process the reference image feature $\mathbf{F}^s$ to compute the matrix $\mathbf{T}$.
} 
\vspace{-4mm}
\label{fig:transform}
\end{figure}

\Paragraph{Novel View Synthesis.}
Given a set of images for a scene, novel view synthesis aims to generate high-quality images at arbitrary viewpoints.
It can be categorized by the number of input images that cover the scene.
One line of work takes as input a single image or stereo images.
These methods use multi-plane images~\cite{srinivasan19,single_view_mpi,Wizadwongsa2021NeX,zhou2018stereo}, layer depth image~\cite{Kopf2020,Shih3DP20}, or point cloud~\cite{Niklaus_TOG_2019,wiles2020synsin} representations to synthesize images at novel views near the input views, \eg 3D photo.
To enable the image synthesis at \emph{arbitrary} novel views, several recent frameworks take hundreds of input images of a scene as the input.
These frameworks leverage different 3D representations to accomplish the task.
Image-based rendering approaches~\cite{Riegler2020FVS,Riegler2020SVS} compute 3D proxy geometry of the scene, and generate images by warping the input frames to the desired novel views.
Neural radiance field schemes~\cite{liu2020neural,nerf,yu2020pixelnerf,kaizhang2020} use  multi-layer perceptrons to implicitly encode the scene for novel view synthesis.
Point cloud-based methods \cite{Meshry_2019_CVPR,npbg2020} solve different optimization problems to construct the point cloud for a specific 3D scene.
Different from these frameworks, our goal is to generate \emph{stylized} novel view images of the 3D scene.
As shown in \figref{motivation}, while existing algorithms can be used for the 3D scene stylization task, they fail to generate high-quality novel view synthesis results with the desired style.

\Paragraph{Image and Video Stylization.}
Image stylization~\cite{Gatys_2016_CVPR} aims to transfer the style of a reference image to the single input image.
Existing methods~\cite{Chen_2017_CVPR,Johnson2016Perceptual,Li_2017_CVPR,Shen_2018_CVPR,ulyanov16} are designed based on feed-forward networks for transferring a set of \emph{pre-defined} styles. 
For \emph{arbitrary} image style transfer, Huang and Belongie~\cite{huang2017adain} use first-order statistics to encode the style information, and transform the image style via the AdaIN normalization layers.
%
%On the other hand, 
The WCT \cite{WCT-NIPS-2017} approach uses whitening and coloring transformation to match the second-order statistics of the input image to those of the reference image.
In addition, the LST \cite{li2018learning} scheme leverages the convolutional neural networks to reduce the computational cost of solving the transformation matrix in the WCT method for real-time universal style transfer.
Most recently, the TPFR method \cite{Svoboda_2020_CVPR} proposes a regularization layer to facilitate the generalization of image stylization models.
%
%\textcolor{red}{Few approaches extend stylization to multple view images such as narrow-baseline stereo images \cite{stereo1,stereo2}, light field images \cite{Hart_2020_WACV}, or views of single objects \cite{kato2018neural}.
%
%In contrast, our method stylizes complex 3D scenes, and produces results at arbitrary viewpoints.}

Video stylization aims to transfer the style of a reference image to a sequence of video frames.
To address the temporal flickering issue produced by the image stylization approaches, numerous approaches~\cite{Chen_2017_ICCV,Chen2020OpticalFD,Gao2018ReCoNetRC,Gupta_2017_ICCV,Huang_2017_CVPR} incorporate optical flow modules to train feed-forward networks for transferring a particular style to the videos.
Several recent frameworks~\cite{deng2020arbitrary,gao2020,Compound2020} enable the video style transfer to \emph{arbitrary} styles.
Although significant advances have been made, existing methods are designed specifically for transferring the style of 2D images or video sequences.
As shown in \figref{motivation}, simply applying these schemes for the 3D scene stylization task leads to problematic results, such as blurry or short/long-range inconsistent images across different novel views.

Several efforts have been made to perform the stylization in 3D space.
However, these approaches are only applicable to single objects~\cite{kato2018neural}, narrow-baseline stereo images~\cite{stereo2,stereo1}, or light field images~\cite{Hart_2020_WACV}.
In contrast, our method stylizes complex 3D scenes, and produces consistent results at arbitrary viewpoints.

\Paragraph{Deep neural networks for point clouds.}
Various deep neural network~(DNN)-based models~\cite{Klokov_2017_ICCV, Le_2018_CVPR,li2018sonet, NEURIPS2018_f5f8590c,qi2016pointnet,qi2017pointnetplusplus,Wu_2019_CVPR,saining2018ascn,zhao2019pointweb} that take point clouds as input are widely studied for vision recognition tasks including 3D semantic segmentation~\cite{armeni2017joint}, 3D shape classification or normal estimation~\cite{Zhirong15CVPR}, and 3D object part segmentation~\cite{Yi16}. 
Recently, Mallya~\etal~\cite{mallya2020world} proposes a point cloud colorization approach for the video-to-video synthesis task. 
In this work, we propose a DNN-based point cloud transformation model for the 3D scene stylization task.
We note that the PSNet~\cite{Cao_2020_WACV} model aims to transfer the style of the point cloud.
Nevertheless, there are two issues for the PSNet method to be applied to the 3D scene stylization task.
First, it does not support synthesizing high-quality stylized images at novel views, which makes the PSNet framework limited for real-world (\eg AR) applications.
Second, since the PSNet scheme requires the optimization process for each specific scene, it is time-consuming, and fails to handle large-scale scenes in the real-world with more than $60$M points, such as those in the Tanks and Temples dataset~\cite{Knapitsch2017}.
In contrast, we propose a feed-forward point cloud model that is efficient, capable of handling large-scale 3D scenes, and generating images with arbitrary styles at various novel views.

\section{Methodology}
\label{sec:method}

We present the overview of the proposed 3D scene stylization framework in \figref{overview}.
Given a set of $N$ input images $\{\mathbf{I}_n\}^N_{n=1}$ of a static scene, and a reference image $\mathbf{S}$ with the desired style, our goal is to synthesize the image $\mathbf{O}_v$ at the novel view $v$ with the camera pose $(\mathbf{R}_v, t_v)$ and intrinsic $\mathbf{K}_v$.
Specifically, the generated novel view image $\mathbf{O}_v$ needs to 1) match the style of the reference image $\mathbf{S}$ and 2) be consistent for different viewpoints $v$.
To handle such (especially the consistency) requirements, our core idea is to 1) construct a single 3D representation, \ie point cloud, for the holistic scene, and 2) transform the representation to produce not only stylized but also consistent novel view synthesis results.
The proposed approach consists of three steps: point cloud creation, point cloud transformation, and novel view synthesis, described in the following sections.

\subsection{Point Cloud Construction}
\label{sec:3_1}

\Paragraph{Pre-processing.}
Our method leverages camera pose and proxy geometry to construct the 3D point cloud.
Given the input images $\{\mathbf{I}_n\}^N_{n=1}$, we first use a structure-from-motion algorithm~\cite{colmap} to estimate the camera poses $\{\mathbf{R}_n, t_n\}^N_{n=1}$ and intrinsic parameters $\{\mathbf{K_n}\}^N_{n=1}$.
For each image $\mathbf{I}_n$, we use the COLMAP \cite{colmap,mvs} and Delaunay-based reconstruction~\cite{m2,m1} schemes to obtain the depth map $\mathbf{D}_n$ that can appropriately back-project the points from the image plane to the 3D space.

\Paragraph{Feature extraction and back-projection.}
Since our goal is to transform the point cloud representation for the 3D scene stylization purpose, we need the point cloud representation to encode the style information.
Therefore, we use the VGG-19 model~\cite{Simonyan15} pre-trained on the ImageNet~\cite{deng2009imagenet} dataset to extract the relu$3\_1$ feature maps $\{\mathbf{F}^c_n\}^{N}_{n=1}$ of the input images $\{\mathbf{I}_n\}^N_{n=1}$.
The width and height of each feature map is $H$ and $W$.
According to the depth map $\{\mathbf{D}_n\}^N_{n=1}$, we back-project all the points in each feature map to build the 3D point cloud $\{f^c_p\}^{P}_{p=1}$, where $P=NHW$ is the total number of points in the constructed point cloud.

\subsection{Point Cloud Transformation}
\label{sec:3_2}

We model the 3D scene stylization process as a linear transformation~\cite{li2018learning} between the constructed and stylized point clouds.
Intuitively, the goal is to match the covariance statistics of the stylized point clouds and those of the reference image $\mathbf{S}$.
To achieve this, we use the pre-trained VGG-19 network to extract the relu$3\_1$ feature map from the reference image $\mathbf{S}$ as the style feature map $\mathbf{F}^s$.
Given the constructed point cloud $\{f^c_p\}^{P}_{p=1}$, we use a predicted linear transformation matrix $\mathbf{T}$ to compute the modulated point cloud $\{f^d_p\}^{P}_{p=1}$, namely
\begin{equation}
    f^d_p = \mathbf{T}(f^c_p - \bar{f}^c) + \bar{f}^s\hspace{5mm}\forall p \in [1,\cdots,P],
    \label{eq:transformation}
\end{equation}
where $\bar{f}^c$ is the mean of the features in the point cloud $\{f^c_p\}^{P}_{p=1}$, and $\bar{f}^s$ is the mean of the style feature map $\mathbf{F}^s$.

\Paragraph{Linear transformation matrix $\mathbf{T}$.} The transformation matrix $\mathbf{T}$ is computed from the style feature map $\mathbf{F}^s$ and constructed point cloud $\{f^c_p\}^{P}_{p=1}$.
As shown in \figref{transform}, we adopt the strategy similar to the LST~\cite{li2018learning} method that uses the convolution layers, covariance computation, and fully-connected layers to compute the matrix $\mathbf{T}^s$ from the style feature map $\mathbf{F}^s$.
On the other hand, we develop a series of point cloud aggregation modules to process the point cloud $\{f^c_p\}^{P}_{p=1}$, and use the covariance computation followed by the fully-connected layers to calculate the matrix $\mathbf{T}^c$.
Finally, we obtain the transformation matrix $\mathbf{T}=\mathbf{T}^s\mathbf{T}^c$.

\Paragraph{Point cloud aggregation.}
It is challenging to gather the information contained in the constructed point cloud $\{f^c_p\}^{P}_{p=1}$ due to the sparsity and non-uniformity.
We note that the constructed point cloud is non-uniform if the input images cover a particular region of the 3D scene.
In this work, we leverage the set abstraction~\cite{qi2017pointnet++} concept to aggregate the point cloud.
The input {\small$\{f^c_p\}^{P}_{p=1}$} to a point cloud aggregation module is a set of $P$ points with feature dimension $c$, and the output {\small$\{f^{c'}_p\}^{P'}_{p=1}$} is a set of $P'$ points with dimension $c'$.
We first sample a subset of $P'$ points $\{f^c_p\}^{P'}_{p=1}$ using the iterative farthest point sampling algorithm~\cite{gonzalez1985clustering,moenning2003fast}.
Viewing the sampled points as the centroids in the 3D space, we use a radius parameter $r$ to find the nearby points to form a point group.
By using the MLP layers and the max pooling operator to map each point group to a vector, we obtain the aggregated point cloud $\{f^{c'}_p\}^{P'}_{p=1}$.
%
%\vspace{-2mm}
%\begin{equation}
%\small
%\{f^{c'}_p\}^{P'}_{p=1}=\mathrm{MaxPool}(\mathrm{MLP}(\{f^{c}_p\}^{P}_{p=1})). \nonumber
%\vspace{-2mm}
%\end{equation}
%
The output {\small$\{f^{c'}_p\}^{P'}_{p=1}$} is then used as the input for the next module.
We use three point cloud aggregation modules sequentially in our pipeline.

\begin{figure*}[ht!]
\centering
\includegraphics[width=\linewidth]{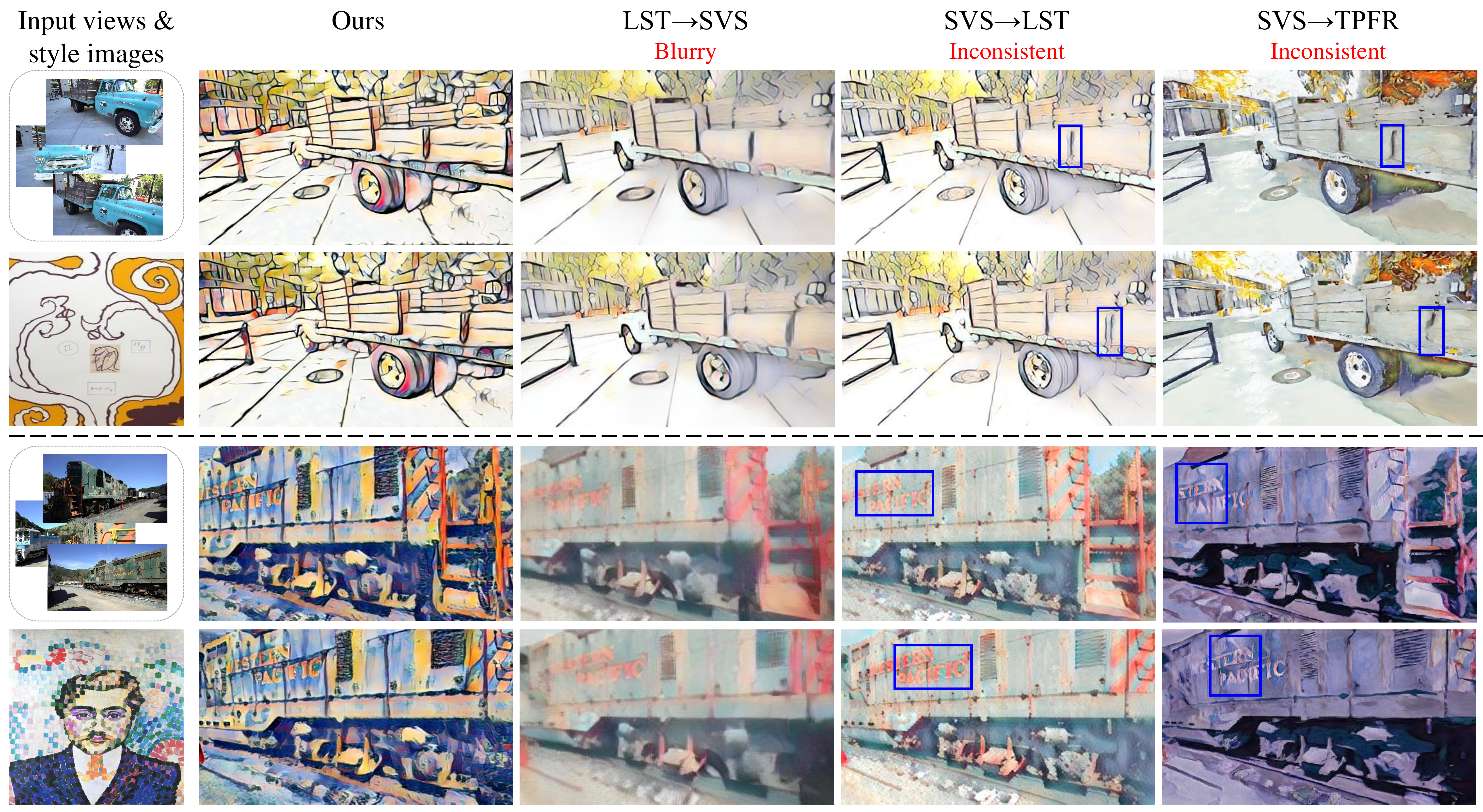}
\caption{
\textbf{Visual comparisons to image stylization-based approaches.} We compare the stylized novel view images generated by the three image stylization alternative schemes and our model on Tanks and Temples dataset \cite{Knapitsch2017}.
}
 
\label{fig:result1}
\end{figure*}

\begin{figure*}[ht!]
\centering
\includegraphics[width=\linewidth]{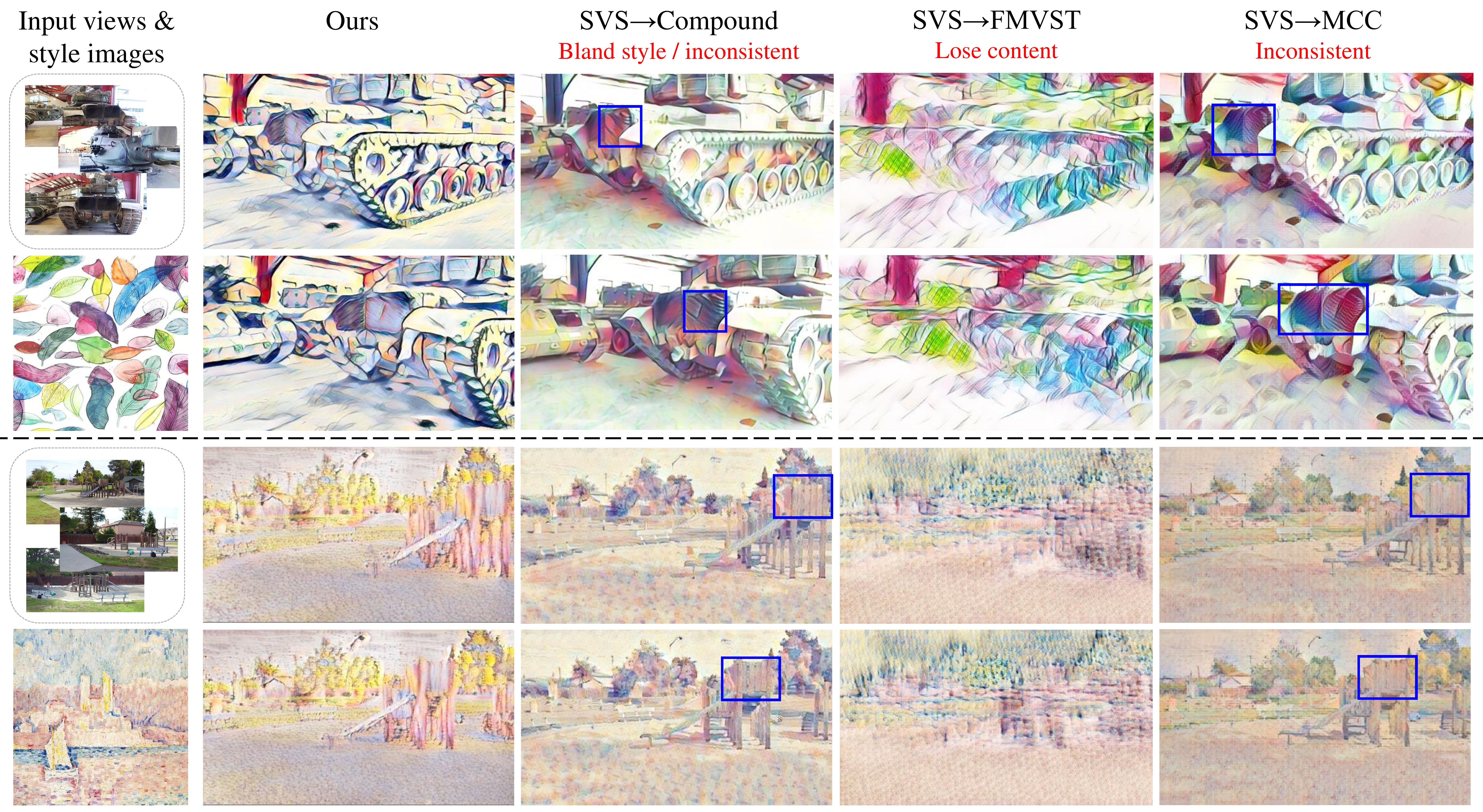}
\caption{
\textbf{Visual comparisons to video stylization-based approaches.} We compare the stylized novel view images generated by the three video stylization alternative schemes and our model on Tanks and Temples dataset~\cite{Knapitsch2017}.
}
\label{fig:result2}
\end{figure*}

\begin{figure*}[t]%[h!]
\centering
\includegraphics[width=0.95\linewidth]{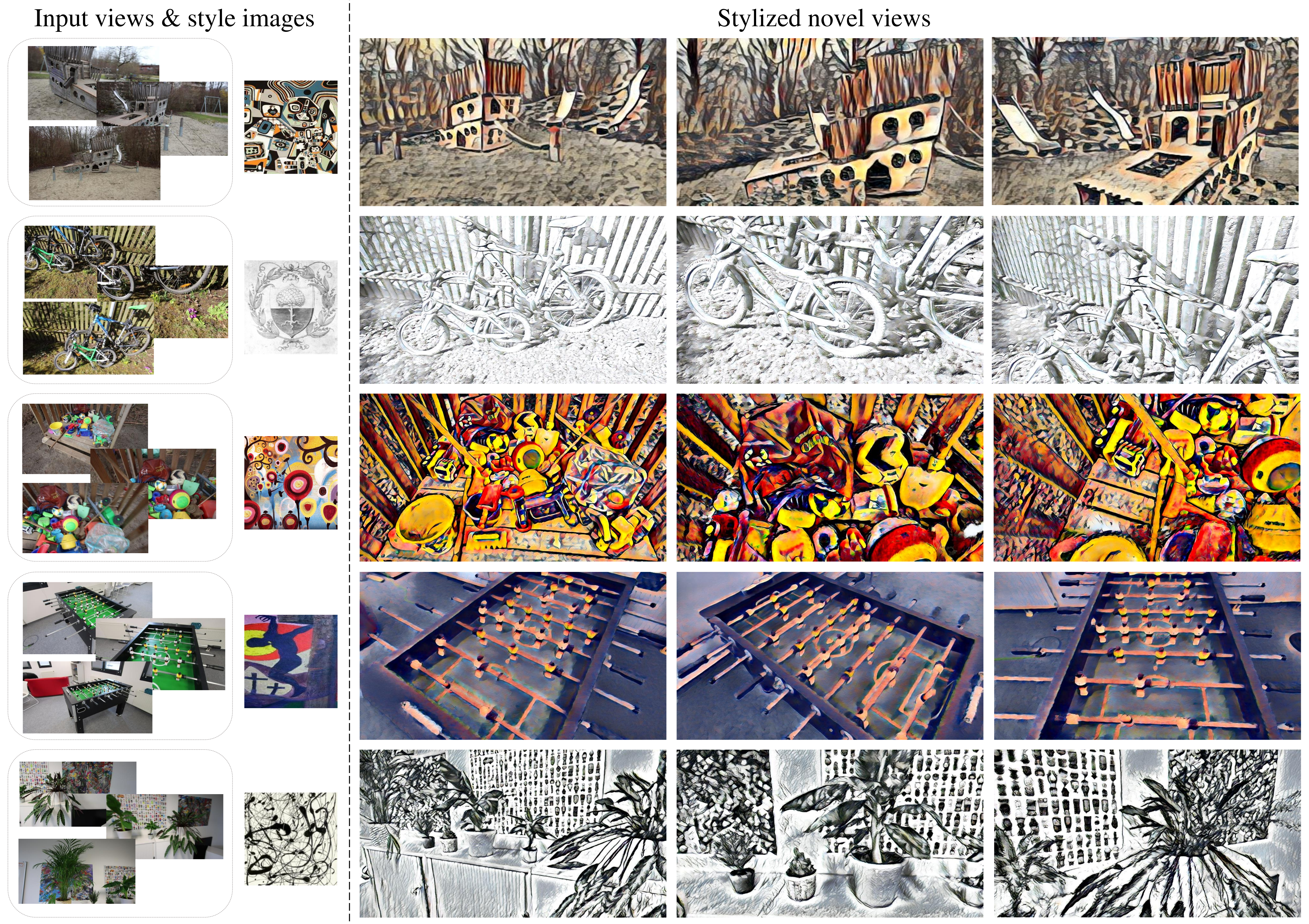}
\caption{
\textbf{Qualitative results on the FVS dataset.} We demonstrate the generalization of the proposed approach by training on the Tanks and Temples dataset, then testing on the FVS dataset.
}
\vspace{-1.5mm}
\label{fig:record}
\end{figure*}

\subsection{Novel View Synthesis and Model Training}
\label{sec:3_3}
We aim to synthesize stylized image $\mathbf{O}_v$ at an arbitrary novel view $v$.
Given the target camera pose $(\mathbf{R}_v, t_v)$ and intrinsic $\mathbf{K}_v$, we use Pytorch3D~\cite{ravi2020pytorch3d,wiles2020synsin} to render the transformed 2D feature map $\mathbf{F}^d_v$.
We then use a decoder network to generate the stylized novel view image $\mathbf{O}_v$ from the 2D feature map $\mathbf{F}^d_v$.

\Paragraph{Model training.} We keep the pre-trained VGG-19 feature extractor fixed during the whole training phase.
We first train the decoder network to perform the non-stylized novel view synthesis.
Since the ground-truth (non-stylized) novel view image is available in the training sets, we use the $\ell1$ reconstruction loss to optimize the decoder network.
We then keep the decoder network fixed, and train the proposed point cloud transformation module with the following loss functions:
\begin{compactitem}
\item{\textbf{Content loss} $\mathcal{L}_c$} ensures the preservation of the content information by measuring the distance between the pre-trained VGG-19 features of the generated stylized image $\mathbf{O}_v$ and the ground-truth (non-stylized) image $\mathbf{I}_v$.

\item{\textbf{Style loss} $\mathcal{L}_s$} encourages the synthesized image $\mathbf{O}_v$ to match the style of the reference image $\mathbf{S}$. 
Similar to recent style transfer approaches~\cite{Johnson2016Perceptual,li2018learning}, we extract the features at different layers of the pre-trained VGG-19 model, and compute the gram matrix differences.
\end{compactitem}
The overall loss function for training the point cloud transformation module is 
\begin{equation}
    \mathcal{L} = \mathcal{L}_c(\mathbf{O}_v, \mathbf{I}_v) + \lambda \mathcal{L}_s(\mathbf{O}_v,\mathbf{S}),
\end{equation}
where $\lambda$ controls the importance of each loss term.
\begin{figure}[t]%[h!]
\centering
\includegraphics[width=0.85\linewidth]{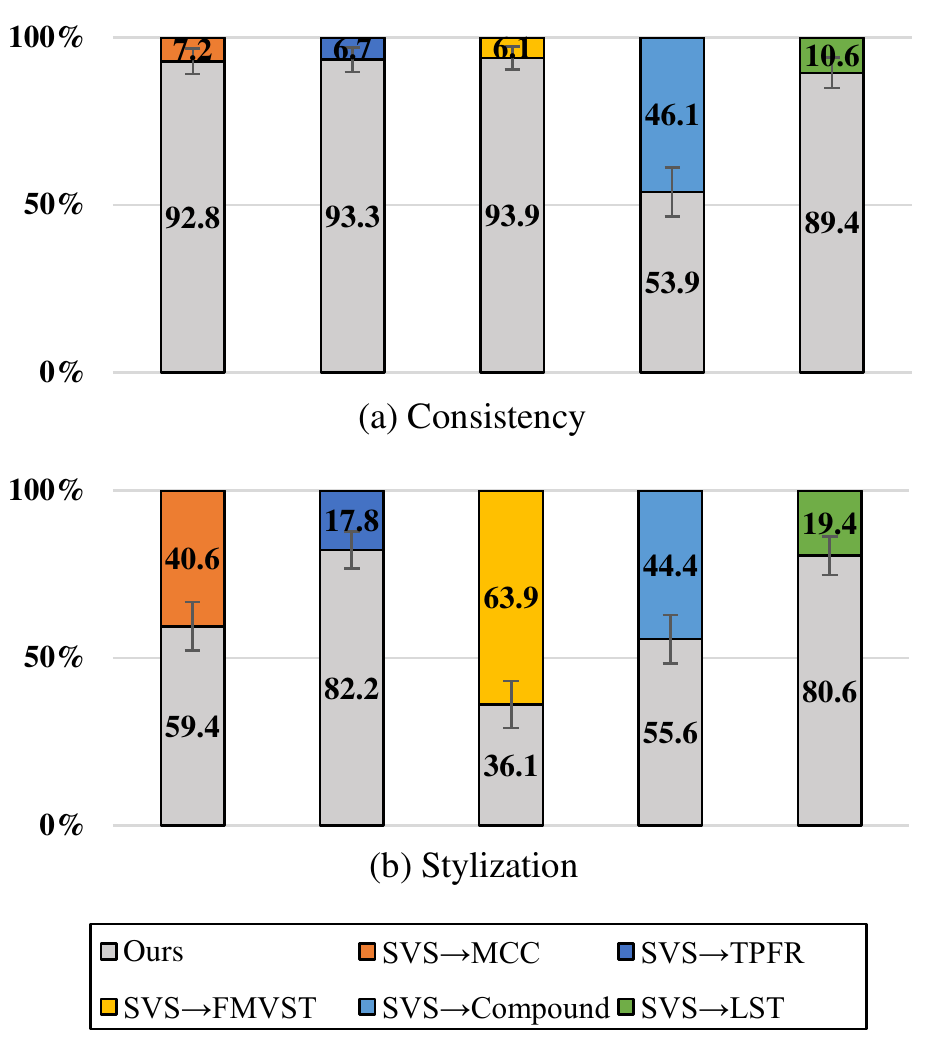}
\caption{
\textbf{User preference study.} We conduct a user study and ask subjects to select the results that (a) have more consistent
contents across different video frames (\eg less flickering), (b) better match the style of the example image. 
The number indicates the percentage of preference.
} 
\vspace{-1.5mm}
\label{fig:user}
\end{figure}

\section{Experimental Results}
\label{sec:results}
We conduct extensive experiments on two real-world datasets to validate the efficacy of the proposed 3D scene stylization model.
%
%More experimental results including videos and implementation details are presented in the supplementary materials.
%
%We will release the source code and the trained model for reproducible research.

\Paragraph{Datasets.}
We use the Tanks and Temples~\cite{Knapitsch2017} dataset for quantitative evaluation.
Similar to the setting in FVS~\cite{Riegler2020FVS}, we use 17 out of the 21 scenes for the training. 
The four remaining scenes (Truck, Train, M60 and Playground) are used for testing.
We also present qualitative results on the FVS~\cite{Riegler2020FVS} dataset, which consists of 6 scenes: Bike, Flowers, Pirate, Digger, Sandbox and Soccertable.
Note that both datasets are collected by \emph{handheld} cameras in un-constraint motions.
%\textcolor{red}{Both datasets are collected by \emph{handheld} cameras in un-constraint motions. We further evaluate our method on forward-facing scenes in LLFF \cite{mildenhall2019llff} and Shiny \cite{Wizadwongsa2021NeX} datasets.}

\Paragraph{Evaluated methods.}
As the 3D scene stylization task is a relatively new problem, we evaluate our method against alternative approaches built upon the state-of-the-art novel view synthesis NeRF++~\cite{kaizhang2020}, SVS~\cite{Riegler2020SVS}, and image/video stylization schemes:
\begin{compactitem}
\item{\textbf{Image stylization $\rightarrow$ novel view synthesis:}} We first use image stylization schemes LST~\cite{li2018learning} or TPFR~\cite{Svoboda_2020_CVPR} to transfer the style to the input images $\{\mathbf{I}_n\}^N_{n=1}$, then perform novel view synthesis.

\item{\textbf{Novel view synthesis $\rightarrow$ image stylization:}} We apply image stylization to the novel view synthesis results.

\item{\textbf{Novel view synthesis $\rightarrow$ video stylization:}} We use a series of novel view synthesis results to create a \emph{video}, then apply video stylization methods Compound~\cite{Compound2020}, FMVST~\cite{gao2020}, or MCC \cite{deng2020arbitrary}.
\end{compactitem}

\subsection{Qualitative Results}
\label{sec:qualitative}
\Paragraph{Image stylization.} \figref{result1} presents the qualitative comparison between the stylized novel view images generated by the three image stylization alternative schemes and the proposed method.
Since the images are stylized independently without considering the consistency issue across different viewpoints, we observe two issues in the image stylization-based methods.
First, LST $\rightarrow$ SVS generally produces blurry novel view images.
Since the stylized input images are not consistent, the novel view synthesis approach tends to \emph{blend} such inconsistency, which leads to blurry results.
Second, the novel view synthesis results are not consistent if we operate in the reverse order, \ie SVS $\rightarrow$ LST.
We highlight the inconsistency using yellow boxes in \figref{result1}.
Note that we observe the same problem if we replace SVS with NeRF++.
%, as demonstrated in the supplementary materials.
%supplementary materials.

\Paragraph{Video stylization.}
We qualitatively evaluate the results by the proposed method and three video stylization alternative approaches in \figref{result2}.
Specifically, we create the videos using a series of novel view synthesis results.
All the alternative approaches generate inconsistent results between two relatively far-away viewpoints since the video stylization methods only guarantee \emph{short-term} consistency in the video.
Although SVS$\rightarrow$Compound generates less inconsistent results, the style of the novel view images is bland and not aligned with that of the reference image.
On the other hand, SVS$\rightarrow$FMVST creates images that better match the desired style, but fails to preserve the content of the original scene.

In contrast to the image and video stylization alternative approaches, our method 1) generates sharp novel view images with correct scene contents and the desired style, and 2) guarantees the short/long-range consistency.
Furthermore, we demonstrate the generalization of the proposed framework in \figref{record} %and \figref{additional}
, where we use the model trained on the Tanks and Temples dataset to perform the 3D scene stylization task on the FVS dataset.%, LLFF, and Shiny dataset.
%
%Please find more qualitative results (\textbf{videos}) in the supplementary materials.

\subsection{Quantitative Results}
\label{sec:quant}
\Paragraph{Stylization quality.}
We conduct a user study to understand the user preference between the proposed and the alternative approaches.
For each testing scene in the Tanks and Temples dataset, we create a video using a series of stylized novel view synthesis results.
By presenting two videos generated by different methods for the same scene, we ask the participants to select the one that (1) has more consistent contents across different video frames (e.g., less flickering), and (2) better matches the style of the reference image.
As the results shown in \figref{user}, the synthesized images by the proposed method are consistent and close to the reference style.
We observe that the users slightly prefer the style generated by SVS$\rightarrow$FMVST.
However, as illustrated in \secref{qualitative} and \figref{result2}, SVS$\rightarrow$FMVST fails to preserve the content of the original scene.

\begin{table}[t]
    \centering
    \scriptsize
	\caption{
	\textbf{Short-range consistency.} We compare the long-range consistency using the warping error ($\downarrow$) between the viewpoints of $(t-1)$-th and $t$-th testing video frames in the Tanks and Temples dataset~\cite{Knapitsch2017}. We report the average errors of 15 diverse styles. The best performance is in \first{bold} and the second best is \second{underscored}.
	}

\begin{tabular}{@{}l|cccc|c@{}} 
    \toprule
    Method & Truck & Playground & Train & M60 & Average \\
    \midrule
NeRF++$\rightarrow$LST	    & 0.215	& 0.168	& 0.250	& 0.274	& 0.231 \\
SVS$\rightarrow$LST	        & 0.192	& 0.159	& 0.220	& 0.241	& 0.206 \\
NeRF++$\rightarrow$TPFR	    & 0.216	& 0.214	& 0.299	& 0.279	& 0.258 \\
SVS$\rightarrow$TPFR	    & 0.235	& 0.237	& 0.291	& 0.276	& 0.264 \\
    \midrule
NeRF++$\rightarrow$Compound	    & 0.188	& 0.169	& 0.229	& 0.208	& 0.202 \\
SVS$\rightarrow$Compound	    & \first{0.166}	& \first{0.156}	& \second{0.199}	& \first{0.160}	& \second{0.172} \\
NeRF++$\rightarrow$FMVST	& 0.342	& 0.300	& 0.405	& 0.348	& 0.354 \\
SVS$\rightarrow$FMVST	    & 0.343	& 0.304	& 0.412	& 0.337	& 0.354 \\
NeRF++$\rightarrow$MCC	    & 0.250	& 0.201	& 0.269	& 0.255	& 0.246 \\
SVS$\rightarrow$MCC 	    & 0.242	& 0.198	& 0.260	& 0.224	& 0.232 \\
    \midrule
Ours	        & \second{0.184}	& \second{0.158}	& \first{0.170}	& \second{0.172}	& \first{0.170} \\
    \bottomrule
\end{tabular} 

\label{tab:short}
\end{table}
\begin{table}[t]
    \centering
    \scriptsize
	\caption{
	\textbf{Long-range consistency.} We compare the long-range consistency using the warping error ($\downarrow$) between the viewpoints of $(t-7)$-th and $t$-th testing video frames in the Tanks and Temples dataset~\cite{Knapitsch2017}. We report the average errors of 15 diverse styles. The best performance is in \first{bold} and the second best is \second{underscored}.
	}
\begin{tabular}{@{}l|cccc|c@{}} 
    \toprule
    Method & Truck & Playground & Train & M60 & Average \\
    \midrule
NeRF++$\rightarrow$LST    & 0.570	& 0.349	& 0.520	& 0.639	& 0.521 \\
SVS$\rightarrow$LST 	        & \second{0.567}	& \first{0.327}	& 0.470	& 0.603	& 0.489 \\
NeRF++$\rightarrow$TPFR	    & 0.579	& 0.436	& 0.503	& 0.655	& 0.541 \\
SVS$\rightarrow$TPFR	    & 0.605	& 0.430	& 0.470	& 0.581	& 0.513 \\
    \midrule
NeRF++$\rightarrow$Compound	& 0.586	& 0.398	& 0.477	& 0.557	& 0.498 \\
SVS$\rightarrow$Compound	& 0.573	& 0.388	& \second{0.422}	& \second{0.460}	& \second{0.449} \\
NeRF++$\rightarrow$FMVST	& 0.742	& 0.525	& 0.636	& 0.695	& 0.644 \\
SVS$\rightarrow$FMVST	    & 0.732	& 0.519	& 0.620	& 0.662	& 0.626 \\
NeRF++$\rightarrow$MCC	    & 0.691	& 0.450	& 0.535	& 0.646	& 0.571 \\
SVS$\rightarrow$MCC	        & 0.693	& 0.447	& 0.516	& 0.584	& 0.548 \\
    \midrule
Ours	        & \first{0.559}	& \second{0.337}	& \first{0.412}	& \first{0.458}	& \first{0.431} \\
    \bottomrule
\end{tabular} 
\vspace{-1mm}
\label{tab:long}
\end{table}
\begin{figure*}[ht!]
\centering
\includegraphics[width=0.95\linewidth]{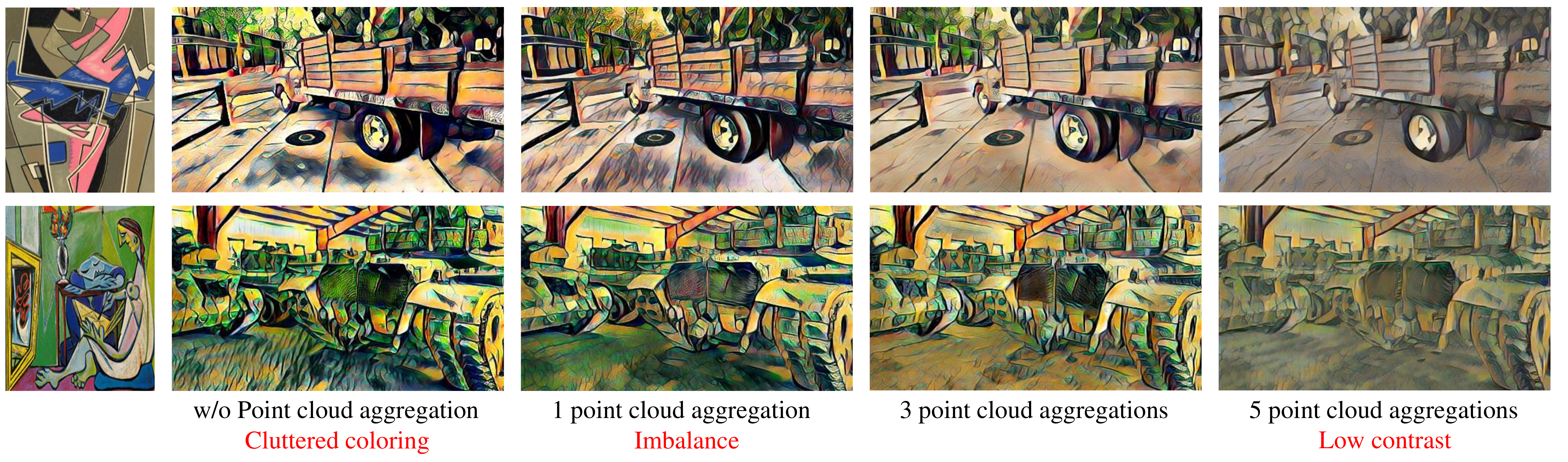}
\caption{
\textbf{Ablation study on the number of point cloud aggregation modules.} We compare the visual results of using $0$/$1$/$\mathbf{3}$/$5$ modules.
We empirically decide to use $3$ modules for better visual quality.
}
\vspace{-2.5mm}
\label{fig:conv}
\end{figure*}

\begin{figure}[t]
\centering
\includegraphics[width=0.77\linewidth]{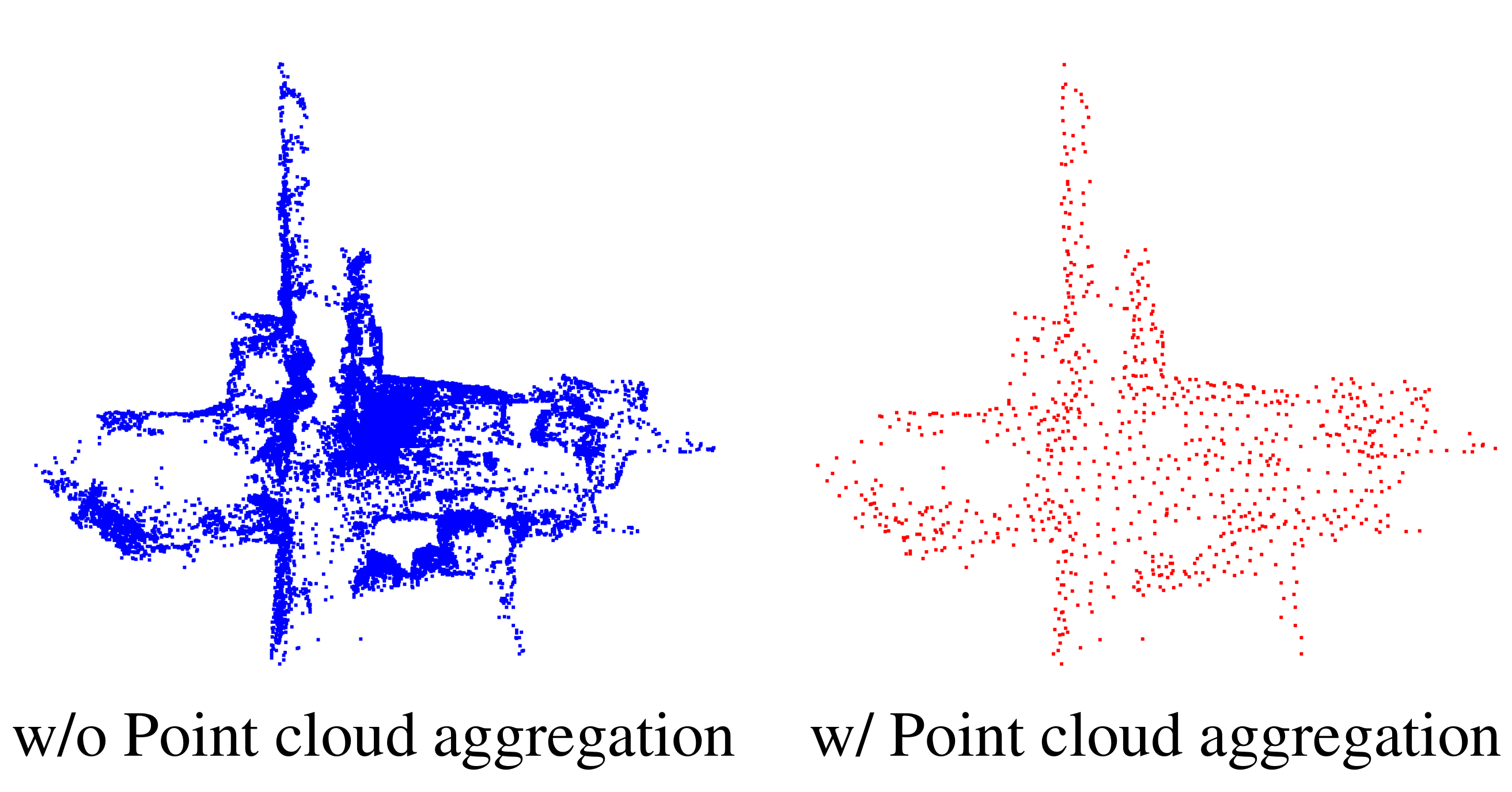}
\caption{
\textbf{Role of point aggregation.} We visualize the point distribution before and after the point aggregation. 
Our point aggregation module obtains a more uniform-distributed point set to fairly estimate the transformation matrix $\mathbf{T}$ that achieves better 3D scene stylization results.}
\vspace{-3.0mm}
\label{fig:point}
\end{figure}

\Paragraph{Short-range consistency.}
We use the warped LPIPS metric~\cite{lpips} to measure the consistency of the results across different viewpoints.
Given a stylized image at a novel view $v$, we warp the results generated at another novel view $v'$ to the view $v$ according to the 3D proxy geometry described in \secref{3_1}.
We then compute the score by
\begin{equation}
    E_\mathrm{warp}(\mathbf{O}_v, \mathbf{O}_v') = \mathrm{LPIPS}(\mathbf{O}_v, W(\mathbf{O}_v'), \mathbf{M}_{v'v}),
\end{equation}
where $W$ is the warping function and $\mathbf{M}_{v'v}$ is the mask of valid pixels warped from the views $v'$ to $v$.
Note that we only use the values of valid pixels in the mask for the ``spatial average" operation in~\cite{lpips}.
For each of the testing scenes in the Tanks and Temples dataset, we use $15$ style images~\cite{gao2020} to compute the average warping error.

We first present the short-range consistency comparison in \tabref{short}.
In this experiment, we use the nearby view for a specific novel view to compute the warping error.\footnote{We use the viewpoints of $(t-1)$-th and $t$-th testing video frames as the views $v'$ and $v$, respectively.}
In general, the image stylization alternative methods produce short-range inconsistent results as they process each novel view independently.
In contrast, the proposed method performs comparably against the video stylization-based approach SVS$\rightarrow$Compound that considers the short-term consistency in videos.
Nevertheless, SVS$\rightarrow$Compound synthesizes bland styles that do not match the desired styles, as the results demonstrated in \figref{result2} and \figref{user}.

\Paragraph{Long-range consistency.}
We also consider the long-range consistency issue in our experiments.
In this experiment, we compute the warping error between the results of two (relatively) far-away views.\footnote{We use the viewpoints of $(t-7)$-th and $t$-th testing video frames as the views $v'$ and $v$, respectively.}
As demonstrated in \tabref{long}, the proposed method performs favorably against the alternative approaches.
Despite the capability of ensuring short-range consistency, video stylization-based schemes fail to maintain the long-range consistency.

\Paragraph{Number of point cloud aggregation modules.}
We conduct an ablation study to decide the number of point cloud aggregation modules described in \secref{3_2}.
The results are presented in \figref{conv}.
We empirically choose to use three modules for better visual quality.
Moreover, we visualize the point distributions before and after the aggregation in \figref{point} to understand the role of the aggregation module.
The point density before the aggregation is higher around the regions of the 3D scene where more input images cover.
As a result, the prediction of the transformation matrix $\mathbf{T}$ is dominated by such regions, which leads to low-quality stylization results (2nd column in \figref{conv}).
By using the point cloud aggregation modules, we obtain a more uniform-distributed point set that fairly estimates the matrix $\mathbf{T}$ for the 3D scene stylization task.

\section{Conclusions}
\label{sec:conclusions}

In this work, we introduce a 3D scene stylization problem that aims to modify the style of the 3D scene and synthesize images at arbitrary novel views. 
We construct a single 3D representation, \ie point cloud, for the holistic scene, and design a point cloud transformation module to transfer the style of the reference image to the 3D representation.
Qualitative and quantitative evaluations validate that our method synthesizes images that 1) contain the desired style and 2) are consistent across various novel views.

\section*{Acknowledgements}
This work is supported in part by the NSF CAREER Grant \#1149783 and a gift from Verisk.

{\small
\bibliographystyle{ieee_fullname}
\bibliography{egbib}

\begin{thebibliography}{100}\itemsep=-1pt

\bibitem{npbg2020}
Kara-Ali Aliev, Artem Sevastopolsky, Maria Kolos, Dmitry Ulyanov, and Victor
  Lempitsky.
\newblock Neural point-based graphics.
\newblock In {\em ECCV}, 2020.

\bibitem{armeni2017joint}
Iro Armeni, Sasha Sax, Amir~R Zamir, and Silvio Savarese.
\newblock Joint 2d-3d-semantic data for indoor scene understanding.
\newblock {\em arXiv preprint arXiv:1702.01105}, 2017.

\bibitem{barron2021mipnerf}
Jonathan~T. Barron, Ben Mildenhall, Matthew Tancik, Peter Hedman, Ricardo
  Martin-Brualla, and Pratul~P. Srinivasan.
\newblock Mip-nerf: A multiscale representation for anti-aliasing neural
  radiance fields.
\newblock In {\em ICCV}, 2021.

\bibitem{boss2020nerd}
Mark Boss, Raphael Braun, Varun Jampani, Jonathan~T. Barron, Ce Liu, and
  Hendrik~P.A. Lensch.
\newblock Nerd: Neural reflectance decomposition from image collections.
\newblock {\em arXiv preprint arXiv:2012.03918}, 2020.

\bibitem{Cao_2020_WACV}
Xu Cao, Weimin Wang, Katashi Nagao, and Ryosuke Nakamura.
\newblock Psnet: A style transfer network for point cloud stylization on
  geometry and color.
\newblock In {\em WACV}, 2020.

\bibitem{chanmonteiro2020piGAN}
Eric Chan, Marco Monteiro, Petr Kellnhofer, Jiajun Wu, and Gordon Wetzstein.
\newblock pi-gan: Periodic implicit generative adversarial networks for
  3d-aware image synthesis.
\newblock In {\em CVPR}, 2021.

\bibitem{chen2021mvsnerf}
Anpei Chen, Zexiang Xu, Fuqiang Zhao, Xiaoshuai Zhang, Fanbo Xiang, Jingyi Yu,
  and Hao Su.
\newblock {M}{V}{S}{N}e{R}{F}: {F}ast {G}eneralizable {R}adiance {F}ield
  {R}econstruction from {M}ulti-{V}iew {S}tereo.
\newblock In {\em ICCV}, 2021.

\bibitem{Chen_2017_ICCV}
Dongdong Chen, Jing Liao, Lu Yuan, Nenghai Yu, and Gang Hua.
\newblock Coherent online video style transfer.
\newblock In {\em ICCV}, 2017.

\bibitem{Chen_2017_CVPR}
Dongdong Chen, Lu Yuan, Jing Liao, Nenghai Yu, and Gang Hua.
\newblock Stylebank: An explicit representation for neural image style
  transfer.
\newblock In {\em CVPR}, 2017.

\bibitem{stereo2}
Dongdong Chen, Lu Yuan, Jing Liao, Nenghai Yu, and Gang Hua.
\newblock Stereoscopic neural style transfer.
\newblock In {\em CVPR}, 2018.

\bibitem{Chen2020OpticalFD}
Xinghao Chen, Yiman Zhang, Yunhe Wang, Han Shu, Chunjing Xu, and Chang Xu.
\newblock Optical flow distillation: Towards efficient and stable video style
  transfer.
\newblock In {\em ECCV}, 2020.

\bibitem{deng2009imagenet}
Jia Deng, Wei Dong, Richard Socher, Li-Jia Li, Kai Li, and Li Fei-Fei.
\newblock Imagenet: A large-scale hierarchical image database.
\newblock In {\em CVPR}, 2009.

\bibitem{kangle2021dsnerf}
Kangle Deng, Andrew Liu, Jun-Yan Zhu, and Deva Ramanan.
\newblock Depth-supervised nerf: Fewer views and faster training for free.
\newblock {\em arXiv preprint arXiv:2107.02791}, 2021.

\bibitem{deng2020arbitrary}
Yingying Deng, Fan Tang, Weiming Dong, haibin Huang, Ma chongyang, and
  Changsheng Xu.
\newblock Arbitrary video style transfer via multi-channel correlation.
\newblock In {\em AAAI}, 2021.

\bibitem{devries2021unconstrained}
Terrance DeVries, Miguel~Angel Bautista, Nitish Srivastava, Graham~W. Taylor,
  and Joshua~M. Susskind.
\newblock Unconstrained scene generation with locally conditioned radiance
  fields.
\newblock In {\em ICCV}, 2021.

\bibitem{du2020nerflow}
Yilun Du, Yinan Zhang, Hong-Xing Yu, Joshua~B. Tenenbaum, and Jiajun Wu.
\newblock Neural radiance flow for 4d view synthesis and video processing.
\newblock In {\em ICCV}, 2021.

\bibitem{Gafni_2021_CVPR}
Guy Gafni, Justus Thies, Michael Zollh{\"o}fer, and Matthias Nie{\ss}ner.
\newblock Dynamic neural radiance fields for monocular 4d facial avatar
  reconstruction.
\newblock In {\em CVPR}, 2021.

\bibitem{Gao2018ReCoNetRC}
Chang Gao, Derun Gu, Fangjun Zhang, and Y. Yu.
\newblock Reconet: Real-time coherent video style transfer network.
\newblock In {\em ACCV}, 2018.

\bibitem{Gaoportraitnerf}
Chen Gao, Yichang Shih, Wei-Sheng Lai, Chia-Kai Liang, and Jia-Bin Huang.
\newblock Portrait neural radiance fields from a single image.
\newblock {\em arXiv preprint arXiv:2012.05903}, 2020.

\bibitem{gao2020}
Wei Gao, Yijun Li, Yihang Yin, and Ming-Hsuan Yang.
\newblock Fast video multi-style transfer.
\newblock In {\em WACV}, 2020.

\bibitem{garbin2021fastnerf}
Stephan~J Garbin, Marek Kowalski, Matthew Johnson, Jamie Shotton, and Julien
  Valentin.
\newblock Fastnerf: High-fidelity neural rendering at 200fps.
\newblock {\em arXiv preprint arXiv:2103.10380}, 2021.

\bibitem{Gatys_2016_CVPR}
Leon~A. Gatys, Alexander~S. Ecker, and Matthias Bethge.
\newblock Image style transfer using convolutional neural networks.
\newblock In {\em CVPR}, 2016.

\bibitem{stereo1}
Xinyu Gong, Haozhi Huang, Lin Ma, Fumin Shen, Wei Liu, and Tong Zhang.
\newblock Neural stereoscopic image style transfer.
\newblock In {\em ECCV}, 2018.

\bibitem{gonzalez1985clustering}
Teofilo~F Gonzalez.
\newblock Clustering to minimize the maximum intercluster distance.
\newblock {\em Theoretical computer science}, 38:293--306, 1985.

\bibitem{guo2020osf}
Michelle Guo, Alireza Fathi, Jiajun Wu, and Thomas Funkhouser.
\newblock Object-centric neural scene rendering.
\newblock {\em arXiv preprint arXiv:2012.08503}, 2020.

\bibitem{Gupta_2017_ICCV}
Agrim Gupta, Justin Johnson, Alexandre Alahi, and Li Fei-Fei.
\newblock Characterizing and improving stability in neural style transfer.
\newblock In {\em ICCV}, 2017.

\bibitem{habtegebrial2020generative}
Tewodros Habtegebrial, Varun Jampani, Orazio Gallo, and Didier Stricker.
\newblock Generative view synthesis: From single-view semantics to novel-view
  images.
\newblock In {\em NeurIPS}, 2020.

\bibitem{Hart_2020_WACV}
David Hart, Bryan Morse, and Jessica Greenland.
\newblock Style transfer for light field photography.
\newblock In {\em WACV}, 2020.

\bibitem{Huang_2017_CVPR}
Haozhi Huang, Hao Wang, Wenhan Luo, Lin Ma, Wenhao Jiang, Xiaolong Zhu, Zhifeng
  Li, and Wei Liu.
\newblock Real-time neural style transfer for videos.
\newblock In {\em CVPR}, 2017.

\bibitem{huang2020semantic}
Hsin-Ping Huang, Hung-Yu Tseng, Hsin-Ying Lee, and Jia-Bin Huang.
\newblock Semantic view synthesis.
\newblock In {\em ECCV}, 2020.

\bibitem{huang2017adain}
Xun Huang and Serge Belongie.
\newblock Arbitrary style transfer in real-time with adaptive instance
  normalization.
\newblock In {\em ICCV}, 2017.

\bibitem{m2}
Michal Jancosek and Tomas Pajdla.
\newblock Multi-view reconstruction preserving weakly-supported surfaces.
\newblock In {\em CVPR}, 2011.

\bibitem{Johnson2016Perceptual}
Justin Johnson, Alexandre Alahi, and Li Fei-Fei.
\newblock Perceptual losses for real-time style transfer and super-resolution.
\newblock In {\em ECCV}, 2016.

\bibitem{kato2018neural}
Hiroharu Kato, Yoshitaka Ushiku, and Tatsuya Harada.
\newblock Neural 3d mesh renderer.
\newblock In {\em CVPR}, 2018.

\bibitem{adam}
Diederik Kingma and Jimmy Ba.
\newblock Adam: A method for stochastic optimization.
\newblock In {\em ICLR}, 2015.

\bibitem{Klokov_2017_ICCV}
Roman Klokov and Victor Lempitsky.
\newblock Escape from cells: Deep kd-networks for the recognition of 3d point
  cloud models.
\newblock In {\em ICCV}, 2017.

\bibitem{Knapitsch2017}
Arno Knapitsch, Jaesik Park, Qian-Yi Zhou, and Vladlen Koltun.
\newblock Tanks and temples: Benchmarking large-scale scene reconstruction.
\newblock {\em ACM TOG (Proc. SIGGRAPH)}, 36(4), 2017.

\bibitem{m1}
Johannes Kopf, Michael Cohen, and Richard Szeliski.
\newblock First-person hyper-lapse videos.
\newblock {\em ACM TOG (Proc. SIGGRAPH)}, 33:1--10, 07 2014.

\bibitem{Kopf2020}
Johannes Kopf, Kevin Matzen, Suhib Alsisan, Ocean Quigley, Francis Ge, Yangming
  Chong, Josh Patterson, Jan-Michael Frahm, Shu Wu, Matthew Yu, et~al.
\newblock One shot 3d photography.
\newblock {\em ACM TOG (Proc. SIGGRAPH)}, 39(4):76--1, 2020.

\bibitem{kosiorek2021nerfvae}
Adam~R. Kosiorek, Heiko Strathmann, Daniel Zoran, Pol Moreno, Rosalia
  Schneider, Soňa Mokrá, and Danilo~J. Rezende.
\newblock Nerf-vae: A geometry aware 3d scene generative model.
\newblock In {\em ICML}, 2021.

\bibitem{Le_2018_CVPR}
Truc Le and Ye Duan.
\newblock Pointgrid: A deep network for 3d shape understanding.
\newblock In {\em CVPR}, 2018.

\bibitem{li2018sonet}
Jiaxin Li, Ben~M Chen, and Gim~Hee Lee.
\newblock So-net: Self-organizing network for point cloud analysis.
\newblock In {\em CVPR}, 2018.

\bibitem{li2021neural3dvideo}
Tianye Li, Mira Slavcheva, M. Zollh{\"o}fer, S. Green, Christoph Lassner,
  Changil Kim, Tanner Schmidt, S. Lovegrove, M. Goesele, and Z. Lv.
\newblock Neural 3d video synthesis.
\newblock {\em arXiv preprint arXiv:2103.02597}, 2021.

\bibitem{li2018learning}
Xueting Li, Sifei Liu, Jan Kautz, and Ming-Hsuan Yang.
\newblock Learning linear transformations for fast arbitrary style transfer.
\newblock In {\em CVPR}, 2019.

\bibitem{NEURIPS2018_f5f8590c}
Yangyan Li, Rui Bu, Mingchao Sun, Wei Wu, Xinhan Di, and Baoquan Chen.
\newblock Pointcnn: Convolution on x-transformed points.
\newblock In {\em NIPS}, 2018.

\bibitem{Li_2017_CVPR}
Yijun Li, Chen Fang, Jimei Yang, Zhaowen Wang, Xin Lu, and Ming-Hsuan Yang.
\newblock Diversified texture synthesis with feed-forward networks.
\newblock In {\em CVPR}, 2017.

\bibitem{WCT-NIPS-2017}
Yijun Li, Chen Fang, Jimei Yang, Zhaowen Wang, Xin Lu, and Ming-Hsuan Yang.
\newblock Universal style transfer via feature transforms.
\newblock In {\em NIPS}, 2017.

\bibitem{li2020neural}
Zhengqi Li, Simon Niklaus, Noah Snavely, and Oliver Wang.
\newblock Neural scene flow fields for space-time view synthesis of dynamic
  scenes.
\newblock In {\em CVPR}, 2021.

\bibitem{autoint2021}
David~B. Lindell, Julien N.~P. Martel, and Gordon Wetzstein.
\newblock Autoint: Automatic integration for fast neural volume rendering.
\newblock In {\em CVPR}, 2021.

\bibitem{liu2020infinite}
Andrew Liu, Richard Tucker, Varun Jampani, Ameesh Makadia, Noah Snavely, and
  Angjoo Kanazawa.
\newblock Infinite nature: Perpetual view generation of natural scenes from a
  single image.
\newblock {\em arXiv preprint arXiv:2012.09855}, 2020.

\bibitem{liu2020neural}
Lingjie Liu, Jiatao Gu, Kyaw~Zaw Lin, Tat-Seng Chua, and Christian Theobalt.
\newblock Neural sparse voxel fields.
\newblock In {\em NeurIPS}, 2020.

\bibitem{liu2021editing}
Steven Liu, Xiuming Zhang, Zhoutong Zhang, Richard Zhang, Jun-Yan Zhu, and
  Bryan Russell.
\newblock Editing conditional radiance fields.
\newblock {\em arXiv preprint arXiv:2105.06466}, 2021.

\bibitem{Lombardi2021MixtureOV}
Stephen Lombardi, Tomas Simon, Gabriel Schwartz, Michael Zollhoefer, Yaser
  Sheikh, and Jason~M. Saragih.
\newblock Mixture of volumetric primitives for efficient neural rendering.
\newblock {\em ACM TOG (Proc. SIGGRAPH)}, 40:1 -- 13, 2021.

\bibitem{mallya2020world}
Arun Mallya, Ting-Chun Wang, Karan Sapra, and Ming-Yu Liu.
\newblock World-consistent video-to-video synthesis.
\newblock In {\em ECCV}, 2020.

\bibitem{martinbrualla2020nerfw}
Ricardo Martin-Brualla, Noha Radwan, Mehdi S.~M. Sajjadi, Jonathan~T. Barron,
  Alexey Dosovitskiy, and Daniel Duckworth.
\newblock {NeRF in the Wild: Neural Radiance Fields for Unconstrained Photo
  Collections}.
\newblock In {\em CVPR}, 2021.

\bibitem{meng2021gnerf}
Quan Meng, Anpei Chen, Haimin Luo, Minye Wu, Hao Su, Lan Xu, Xuming He, and
  Jingyi Yu.
\newblock {G}{N}e{R}{F}: {G}{A}{N}-based {N}eural {R}adiance {F}ield without
  {P}osed {C}amera.
\newblock In {\em ICCV}, 2021.

\bibitem{Meshry_2019_CVPR}
Moustafa Meshry, Dan~B. Goldman, Sameh Khamis, Hugues Hoppe, Rohit Pandey, Noah
  Snavely, and Ricardo Martin-Brualla.
\newblock Neural rerendering in the wild.
\newblock In {\em CVPR}, 2019.

\bibitem{mildenhall2019llff}
Ben Mildenhall, Pratul~P. Srinivasan, Rodrigo Ortiz-Cayon, Nima~Khademi
  Kalantari, Ravi Ramamoorthi, Ren Ng, and Abhishek Kar.
\newblock Local light field fusion: Practical view synthesis with prescriptive
  sampling guidelines.
\newblock {\em ACM TOG (Proc. SIGGRAPH)}, 38:1 -- 14, 2019.

\bibitem{nerf}
Ben Mildenhall, Pratul~P Srinivasan, Matthew Tancik, Jonathan~T Barron, Ravi
  Ramamoorthi, and Ren Ng.
\newblock Nerf: Representing scenes as neural radiance fields for view
  synthesis.
\newblock In {\em ECCV}, 2020.

\bibitem{moenning2003fast}
Carsten Moenning and Neil~A Dodgson.
\newblock Fast marching farthest point sampling.
\newblock Technical report, University of Cambridge, Computer Laboratory, 2003.

\bibitem{neff2021donerf}
Thomas Neff, Pascal Stadlbauer, Mathias Parger, Andreas Kurz, Joerg~H. Mueller,
  Chakravarty R.~Alla Chaitanya, Anton~S. Kaplanyan, and Markus Steinberger.
\newblock {DONeRF: Towards Real-Time Rendering of Compact Neural Radiance
  Fields using Depth Oracle Networks}.
\newblock {\em Computer Graphics Forum}, 40(4), 2021.

\bibitem{niemeyer2021campari}
Michael Niemeyer and Andreas Geiger.
\newblock Campari: Camera-aware decomposed generative neural radiance fields.
\newblock {\em arXiv preprint arXiv:2103.17269}, 2021.

\bibitem{Niemeyer2020GIRAFFE}
Michael Niemeyer and Andreas Geiger.
\newblock Giraffe: Representing scenes as compositional generative neural
  feature fields.
\newblock In {\em CVPR}, 2021.

\bibitem{Niklaus_TOG_2019}
Simon Niklaus, Long Mai, Jimei Yang, and Feng Liu.
\newblock 3d ken burns effect from a single image.
\newblock {\em ACM TOG (Proc. SIGGRAPH)}, 38(6):1--15, 2019.

\bibitem{noguchi2021neural}
Atsuhiro Noguchi, Xiao Sun, Stephen Lin, and Tatsuya Harada.
\newblock Neural articulated radiance field.
\newblock {\em arXiv preprint arXiv:2104.03110}, 2021.

\bibitem{ost2020neuralscenegraphs}
Julian Ost, Fahim Mannan, Nils Thuerey, Julian Knodt, and Felix Heide.
\newblock Neural scene graphs for dynamic scenes.
\newblock In {\em CVPR}, 2021.

\bibitem{park2020nerfies}
Keunhong Park, Utkarsh Sinha, Jonathan~T. Barron, Sofien Bouaziz, Dan~B
  Goldman, Steven~M. Seitz, and Ricardo Martin-Brualla.
\newblock Deformable neural radiance fields.
\newblock In {\em ICCV}, 2021.

\bibitem{park2021hypernerf}
Keunhong Park, Utkarsh Sinha, Peter Hedman, Jonathan~T. Barron, Sofien Bouaziz,
  Dan~B Goldman, Ricardo Martin-Brualla, and Steven~M. Seitz.
\newblock Hypernerf: A higher-dimensional representation for topologically
  varying neural radiance fields.
\newblock {\em arXiv preprint arXiv:2106.13228}, 2021.

\bibitem{peng2021animatable}
Sida Peng, Junting Dong, Qianqian Wang, Shangzhan Zhang, Qing Shuai, Hujun Bao,
  and Xiaowei Zhou.
\newblock Animatable neural radiance fields for human body modeling.
\newblock In {\em ICCV}, 2021.

\bibitem{peng2021neural}
Sida Peng, Yuanqing Zhang, Yinghao Xu, Qianqian Wang, Qing Shuai, Hujun Bao,
  and Xiaowei Zhou.
\newblock Neural body: Implicit neural representations with structured latent
  codes for novel view synthesis of dynamic humans.
\newblock In {\em CVPR}, 2021.

\bibitem{pumarola2020d}
Albert Pumarola, Enric Corona, Gerard Pons-Moll, and Francesc Moreno-Noguer.
\newblock D-nerf: Neural radiance fields for dynamic scenes.
\newblock In {\em CVPR}, 2021.

\bibitem{qi2016pointnet}
Charles~R Qi, Hao Su, Kaichun Mo, and Leonidas~J Guibas.
\newblock Pointnet: Deep learning on point sets for 3d classification and
  segmentation.
\newblock In {\em CVPR}, 2017.

\bibitem{qi2017pointnetplusplus}
Charles~Ruizhongtai Qi, Li Yi, Hao Su, and Leonidas~J Guibas.
\newblock Pointnet++: Deep hierarchical feature learning on point sets in a
  metric space.
\newblock In {\em NIPS}, 2017.

\bibitem{qi2017pointnet++}
Charles~R Qi, Li Yi, Hao Su, and Leonidas~J Guibas.
\newblock Pointnet++: Deep hierarchical feature learning on point sets in a
  metric space.
\newblock In {\em NIPS}, 2017.

\bibitem{raj2021pva}
Amit Raj, Michael Zollhoefer, Tomas Simon, Jason Saragih, Shunsuke Saito, James
  Hays, and Stephen Lombardi.
\newblock Pva: Pixel-aligned volumetric avatars.
\newblock In {\em CVPR}, 2021.

\bibitem{ravi2020pytorch3d}
Nikhila Ravi, Jeremy Reizenstein, David Novotny, Taylor Gordon, Wan-Yen Lo,
  Justin Johnson, and Georgia Gkioxari.
\newblock Accelerating 3d deep learning with pytorch3d.
\newblock {\em arXiv preprint arXiv:2007.08501}, 2020.

\bibitem{Rebain_2021_CVPR}
Daniel Rebain, Wei Jiang, Soroosh Yazdani, Ke Li, Kwang~Moo Yi, and Andrea
  Tagliasacchi.
\newblock Derf: Decomposed radiance fields.
\newblock In {\em CVPR}, 2021.

\bibitem{reiser2021kilonerf}
Christian Reiser, Songyou Peng, Yiyi Liao, and Andreas Geiger.
\newblock Kilonerf: Speeding up neural radiance fields with thousands of tiny
  mlps.
\newblock In {\em ICCV}, 2021.

\bibitem{rematasICML21}
Konstantinos Rematas, Ricardo Martin-Brualla, and Vittorio Ferrari.
\newblock Sharf: Shape-conditioned radiance fields from a single view.
\newblock In {\em ICML}, 2021.

\bibitem{Riegler2020FVS}
Gernot Riegler and Vladlen Koltun.
\newblock Free view synthesis.
\newblock In {\em ECCV}, 2020.

\bibitem{Riegler2020SVS}
Gernot Riegler and Vladlen Koltun.
\newblock Stable view synthesis.
\newblock In {\em CVPR}, 2021.

\bibitem{colmap}
Johannes~L. Schonberger and Jan-Michael Frahm.
\newblock Structure-from-motion revisited.
\newblock In {\em CVPR}, 2016.

\bibitem{Schwarz2020NEURIPS}
Katja Schwarz, Yiyi Liao, Michael Niemeyer, and Andreas Geiger.
\newblock Graf: Generative radiance fields for 3d-aware image synthesis.
\newblock In {\em NeurIPS}, 2020.

\bibitem{mvs}
Johannes Schönberger, Enliang Zheng, Marc Pollefeys, and Jan-Michael Frahm.
\newblock Pixelwise view selection for unstructured multi-view stereo.
\newblock In {\em ECCV}, 2016.

\bibitem{Shen_2018_CVPR}
Falong Shen, Shuicheng Yan, and Gang Zeng.
\newblock Neural style transfer via meta networks.
\newblock In {\em CVPR}, 2018.

\bibitem{Shih3DP20}
Meng-Li Shih, Shih-Yang Su, Johannes Kopf, and Jia-Bin Huang.
\newblock 3d photography using context-aware layered depth inpainting.
\newblock In {\em CVPR}, 2020.

\bibitem{Simonyan15}
Karen Simonyan and Andrew Zisserman.
\newblock Very deep convolutional networks for large-scale image recognition.
\newblock In {\em ICLR}, 2015.

\bibitem{nerv2021}
Pratul~P. Srinivasan, Boyang Deng, Xiuming Zhang, Matthew Tancik, Ben
  Mildenhall, and Jonathan~T. Barron.
\newblock Nerv: Neural reflectance and visibility fields for relighting and
  view synthesis.
\newblock In {\em CVPR}, 2021.

\bibitem{srinivasan19}
Pratul~P. Srinivasan, Richard Tucker, Jonathan~T. Barron, Ravi Ramamoorthi, Ren
  Ng, and Noah Snavely.
\newblock Pushing the boundaries of view extrapolation with multiplane images.
\newblock In {\em CVPR}, 2019.

\bibitem{su2021anerf}
Shih-Yang Su, Frank Yu, Michael Zollhoefer, and Helge Rhodin.
\newblock A-nerf: Surface-free human 3d pose refinement via neural rendering.
\newblock {\em arXiv preprint arXiv:2102.06199}, 2021.

\bibitem{Sucararxiv2021}
Edgar Sucar, Shikun Liu, Joseph Ortiz, and Andrew Davison.
\newblock {iMAP}: Implicit mapping and positioning in real-time.
\newblock {\em arXiv preprint arXiv:2103.12352}, 2021.

\bibitem{Svoboda_2020_CVPR}
Jan Svoboda, Asha Anoosheh, Christian Osendorfer, and Jonathan Masci.
\newblock Two-stage peer-regularized feature recombination for arbitrary image
  style transfer.
\newblock In {\em CVPR}, 2020.

\bibitem{tancik2020meta}
Matthew Tancik, Ben Mildenhall, Terrance Wang, Divi Schmidt, Pratul~P.
  Srinivasan, Jonathan~T. Barron, and Ren Ng.
\newblock Learned initializations for optimizing coordinate-based neural
  representations.
\newblock In {\em CVPR}, 2021.

\bibitem{Tretschk20arxiv_NR}
Edgar Tretschk, Ayush Tewari, Vladislav Golyanik, Michael Zollh{\"o}fer,
  Christoph Lassner, and Christian Theobalt.
\newblock Non-rigid neural radiance fields: Reconstruction and novel view
  synthesis of a deforming scene from monocular video.
\newblock In {\em ICCV}, 2021.

\bibitem{grf2020}
Alex Trevithick and Bo Yang.
\newblock Grf: Learning a general radiance field for 3d scene representation
  and rendering.
\newblock {\em arXiv preprint arXiv:2010.04595}, 2020.

\bibitem{single_view_mpi}
Richard Tucker and Noah Snavely.
\newblock Single-view view synthesis with multiplane images.
\newblock In {\em CVPR}, 2020.

\bibitem{ulyanov16}
Dmitry Ulyanov, Vadim Lebedev, Andrea Vedaldi, and Victor Lempitsky.
\newblock Texture networks: Feed-forward synthesis of textures and stylized
  images.
\newblock In {\em ICML}, 2016.

\bibitem{wang2021ibrnet}
Qianqian Wang, Zhicheng Wang, Kyle Genova, Pratul Srinivasan, Howard Zhou,
  Jonathan~T. Barron, Ricardo Martin-Brualla, Noah Snavely, and Thomas
  Funkhouser.
\newblock Ibrnet: Learning multi-view image-based rendering.
\newblock In {\em CVPR}, 2021.

\bibitem{Compound2020}
Wenjing Wang, Jizheng Xu, Li Zhang, Yue Wang, and Jiaying Liu.
\newblock Consistent video style transfer via compound regularization.
\newblock In {\em AAAI}, 2020.

\bibitem{wang2020learning}
Ziyan Wang, Timur Bagautdinov, Stephen Lombardi, Tomas Simon, Jason Saragih,
  Jessica Hodgins, and Michael Zollhöfer.
\newblock Learning compositional radiance fields of dynamic human heads.
\newblock In {\em CVPR}, 2021.

\bibitem{wang2021nerfmm}
Zirui Wang, Shangzhe Wu, Weidi Xie, Min Chen, and Victor~Adrian Prisacariu.
\newblock Ne{RF}$--$: Neural radiance fields without known camera parameters.
\newblock {\em arXiv preprint arXiv:2102.07064}, 2021.

\bibitem{wiles2020synsin}
Olivia Wiles, Georgia Gkioxari, Richard Szeliski, and Justin Johnson.
\newblock {SynSin}: {E}nd-to-end view synthesis from a single image.
\newblock In {\em CVPR}, 2020.

\bibitem{Wizadwongsa2021NeX}
Suttisak Wizadwongsa, Pakkapon Phongthawee, Jiraphon Yenphraphai, and Supasorn
  Suwajanakorn.
\newblock Nex: Real-time view synthesis with neural basis expansion.
\newblock In {\em CVPR}, 2021.

\bibitem{Wu_2019_CVPR}
Wenxuan Wu, Zhongang Qi, and Li Fuxin.
\newblock Pointconv: Deep convolutional networks on 3d point clouds.
\newblock In {\em CVPR}, 2019.

\bibitem{Zhirong15CVPR}
Zhirong Wu, Shuran Song, Aditya Khosla, Fisher Yu, Linguang Zhang, Xiaoou Tang,
  and Jianxiong Xiao.
\newblock 3d shapenets: A deep representation for volumetric shapes.
\newblock In {\em CVPR}, 2015.

\bibitem{Xian2021}
Wenqi Xian, Jia-Bin Huang, Johannes Kopf, and Changil Kim.
\newblock Space-time neural irradiance fields for free-viewpoint video.
\newblock In {\em CVPR}, 2021.

\bibitem{xie2021fignerf}
Christopher Xie, Keunhong Park, Ricardo Martin-Brualla, and Matthew Brown.
\newblock Fig-nerf: Figure-ground neural radiance fields for 3d object category
  modelling.
\newblock {\em arXiv preprint arXiv:2104.08418}, 2021.

\bibitem{saining2018ascn}
Saining Xie, Sainan Liu, Zeyu Chen, and Zhuowen Tu.
\newblock Attentional shapecontextnet for point cloud recognition.
\newblock In {\em CVPR}, 2018.

\bibitem{yen2020inerf}
Lin Yen-Chen, Pete Florence, Jonathan~T. Barron, Alberto Rodriguez, Phillip
  Isola, and Tsung-Yi Lin.
\newblock {iNeRF}: Inverting neural radiance fields for pose estimation.
\newblock In {\em IROS}, 2021.

\bibitem{Yi16}
Li Yi, Vladimir~G Kim, Duygu Ceylan, I-Chao Shen, Mengyan Yan, Hao Su, Cewu Lu,
  Qixing Huang, Alla Sheffer, and Leonidas Guibas.
\newblock A scalable active framework for region annotation in 3d shape
  collections.
\newblock {\em ACM TOG (Proc. SIGGRAPH)}, 35(6):1--12, 2016.

\bibitem{yu2021plenoctrees}
Alex Yu, Ruilong Li, Matthew Tancik, Hao Li, Ren Ng, and Angjoo Kanazawa.
\newblock Plenoctrees for real-time rendering of neural radiance fields.
\newblock In {\em ICCV}, 2021.

\bibitem{yu2020pixelnerf}
Alex Yu, Vickie Ye, Matthew Tancik, and Angjoo Kanazawa.
\newblock {pixelNeRF}: Neural radiance fields from one or few images.
\newblock In {\em CVPR}, 2021.

\bibitem{kaizhang2020}
Kai Zhang, Gernot Riegler, Noah Snavely, and Vladlen Koltun.
\newblock Nerf++: Analyzing and improving neural radiance fields.
\newblock {\em arXiv preprint arXiv:2010.07492}, 2020.

\bibitem{lpips}
Richard Zhang, Phillip Isola, Alexei~A Efros, Eli Shechtman, and Oliver Wang.
\newblock The unreasonable effectiveness of deep features as a perceptual
  metric.
\newblock In {\em CVPR}, 2018.

\bibitem{nerfactor}
Xiuming Zhang, Pratul~P Srinivasan, Boyang Deng, Paul Debevec, William~T
  Freeman, and Jonathan~T Barron.
\newblock {NeRFactor: Neural Factorization of Shape and Reflectance Under an
  Unknown Illumination}.
\newblock {\em ACM TOG (Proc. SIGGRAPH Asia)}, 2021.

\bibitem{zhao2019pointweb}
Hengshuang Zhao, Li Jiang, Chi-Wing Fu, and Jiaya Jia.
\newblock {PointWeb}: Enhancing local neighborhood features for point cloud
  processing.
\newblock In {\em CVPR}, 2019.

\bibitem{ZhiICCV2021}
Shuaifeng Zhi, Tristan Laidlow, Stefan Leutenegger, and Andrew Davison.
\newblock In-place scene labelling and understanding with implicit scene
  representation.
\newblock In {\em ICCV}, 2021.

\bibitem{zhou2018stereo}
Tinghui Zhou, Richard Tucker, John Flynn, Graham Fyffe, and Noah Snavely.
\newblock Stereo magnification: learning view synthesis using multiplane
  images.
\newblock {\em ACM TOG (Proc. SIGGRAPH)}, 37(4):1--12, 2018.

\end{thebibliography}
}

\clearpage
\onecolumn
\appendix

\section{Supplementary Materials}

\subsection{Overview}
In this supplementary document, we first present additional experimental results, including run-time analysis.
Second, we provide the implementation details of the proposed framework.
Third, we compare our method (\ie explicit representations) and the current approaches based on implicit representations.
Finally, we discuss the limitations of the proposed scheme and the future research directions.
More qualitative comparisons are available at {\small\url{https://hhsinping.github.io/3d_scene_stylization}}.

\subsection{Additional Experimental Results}

\subsubsection{LLFF and Shiny datasets}
To demonstrate the generalization ability of the proposed method, we use the model trained on the Tanks and Temples dataset~\cite{Knapitsch2017} to produce the stylization results on two additional datasets: LLFF~\cite{mildenhall2019llff} and Shiny~\cite{Wizadwongsa2021NeX}.

\begin{figure}[h!]
\centering
\includegraphics[width=0.5\linewidth]{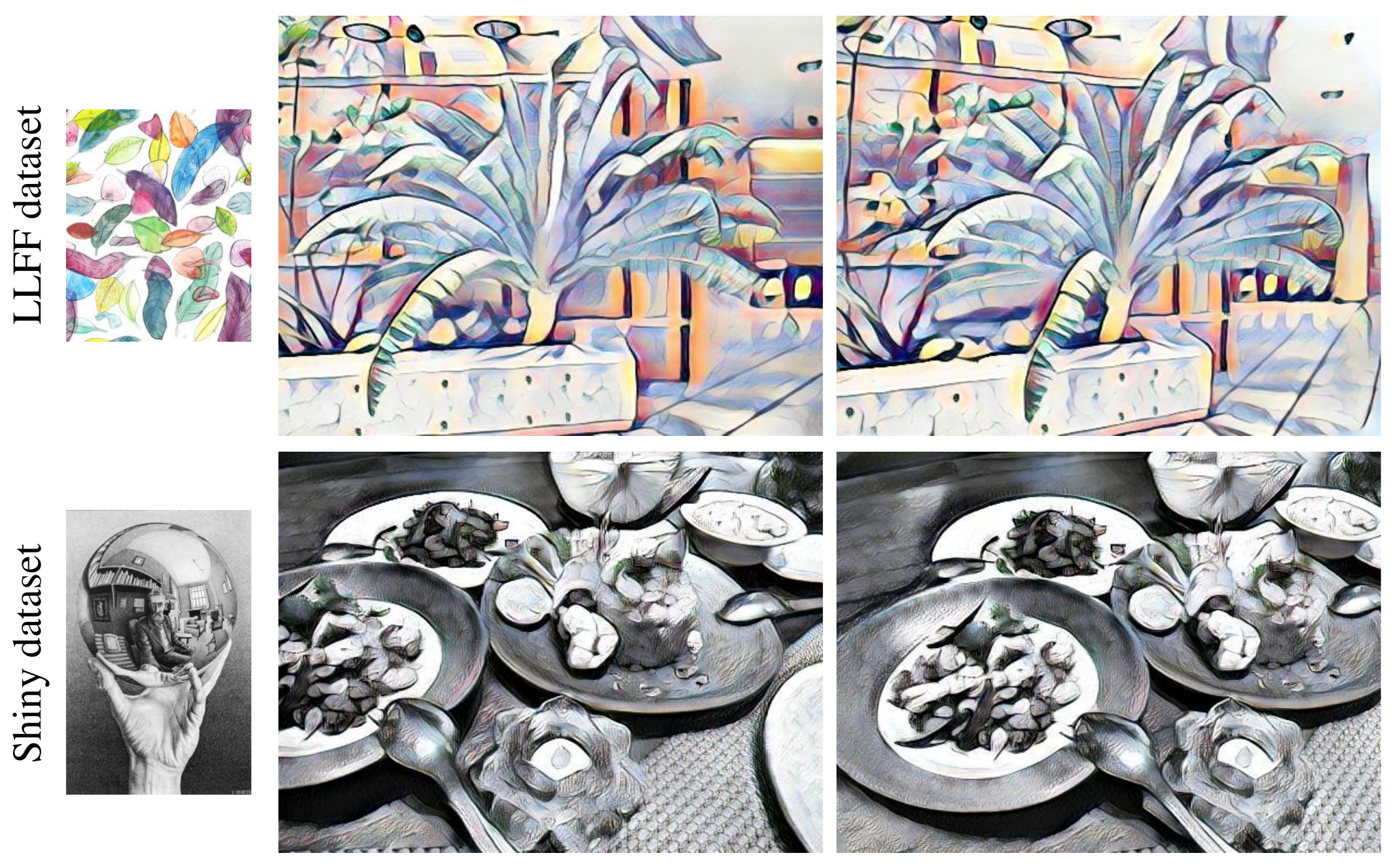}
\caption{
\textbf{Additional results.} We show additional results on LLFF and Shiny datasets using the model trained on the Tanks an Temples dataset.
} 
\vspace{-1.5mm}
\label{fig:additional}
\end{figure}

\subsubsection{LST$\rightarrow$NeRF++}
\begin{figure*}[ht!]
\centering
\includegraphics[width=0.95\linewidth]{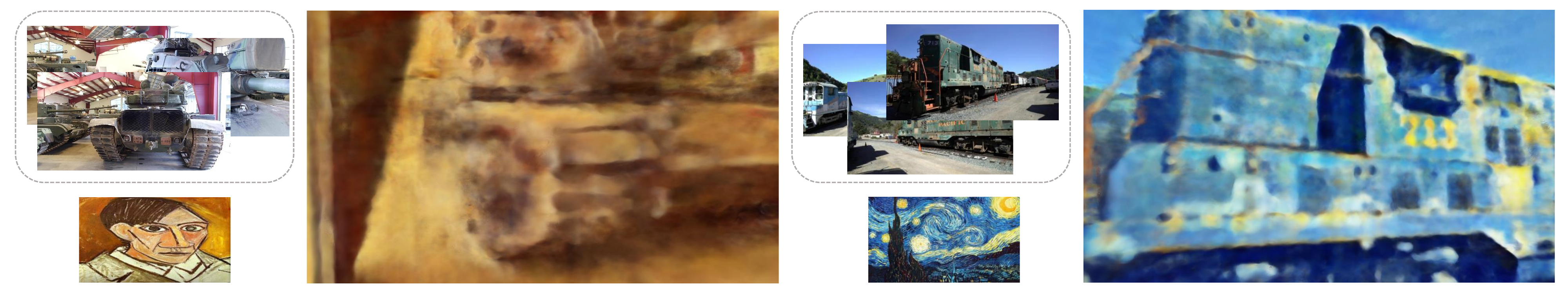}
\vspace{-1.5mm}
\caption{
\textbf{LST$\rightarrow$NeRF++.} The NeRF++ approach produces blurry results if the input images are not consistent due to the per-image stylization by the LST approach.
%
%We present the results of the alternative scheme that first uses LST then applies NeRF++. The results are blurry as the stylized input images are not consistent.
}
\vspace{-3mm}
\label{fig:nerf}
\end{figure*}

As shown in Figure 5 in the paper, applying image stylization schemes before the SVS~\cite{Riegler2020SVS} framework produces blurry results.
In this experiment, we show that replacing the SVS~\cite{Riegler2020SVS} with NeRF++~\cite{kaizhang2020} approaches suffers from the similar issue.
In \figref{nerf}, we present the results LST~\cite{li2018learning} $\rightarrow$ NeRF++~\cite{kaizhang2020}.
Since the input images are not consistent due to the per-image stylization by the LST approach, the NeRF++ model tends to \emph{blend} such inconsistency, which leads to blurry results.

\subsubsection{Ablation Study on Stylization Level}
\begin{figure*}[ht!]
\centering
\includegraphics[width=0.75\linewidth]{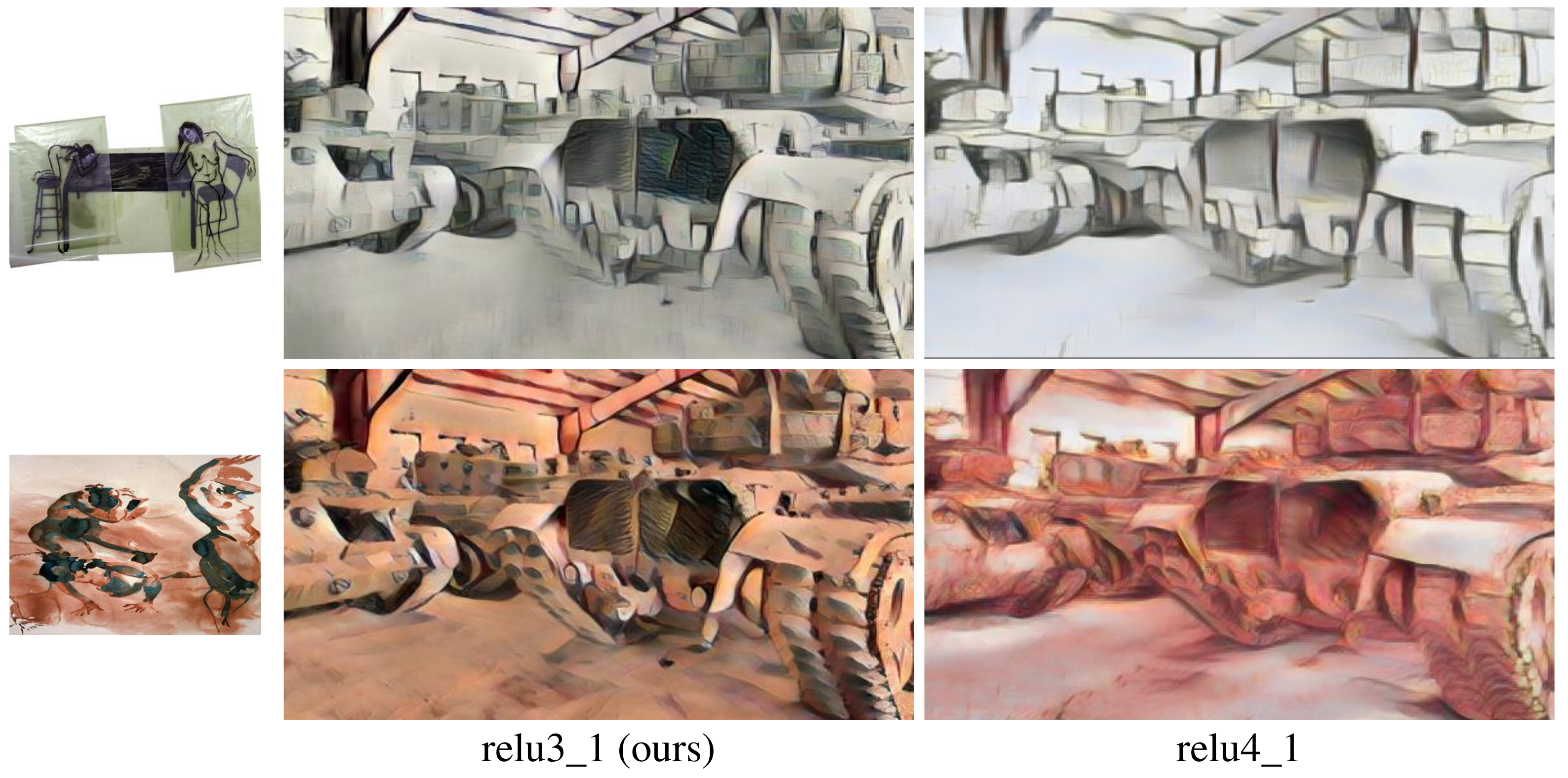}
\vspace{-1.5mm}
\caption{
\textbf{Ablation study on stylization level.} 
We show that our framework is able to transfer styles of different levels.
Extracting the image features from the deeper layers (relu$4\_1$) of the pre-trained VGG-19 network produces more distortion, while using features from the shallower layers (relu$3\_1$) generates more photo-realistic stylization effects.
}
\vspace{-3mm}
\label{fig:level}
\end{figure*}

We use the pre-trained VGG-19 model~\cite{Simonyan15} to extract the feature of the input images for the point cloud construction.
By extracting the features from different layers of the VGG-19 network, our point cloud representation encodes different levels of the style information.
\figref{level} demonstrates that our framework is capable of transferring the different style levels.
Specifically, building the point cloud representation using the deeper (\eg relu$4\_1$) features produces more distortion, while using the shallower (\eg relu$3\_1$) features generates more photo-realistic (\ie preserve more content information) effects.

\subsubsection{Ablation Study on Point Cloud Aggregation}
To gather the style information of the constructed point cloud $\{f^c_p\}^{P}_{p=1}$, we sample a subset of $P'$ points $\{f^c_p\}^{P'}_{p=1}$ and then use a radius parameter $r$ to find $k$ nearby points to form a point group.
Each point group is aggregated to a vector by MLP layers and the max pooling operator to form the aggregated point cloud $\{f^{c'}_p\}^{P'}_{p=1}$.
We conduct the following ablation studies to analyze the hyper-parameters $r$ and $k$.

\begin{figure*}[ht!]
\centering
\includegraphics[width=0.95\linewidth]{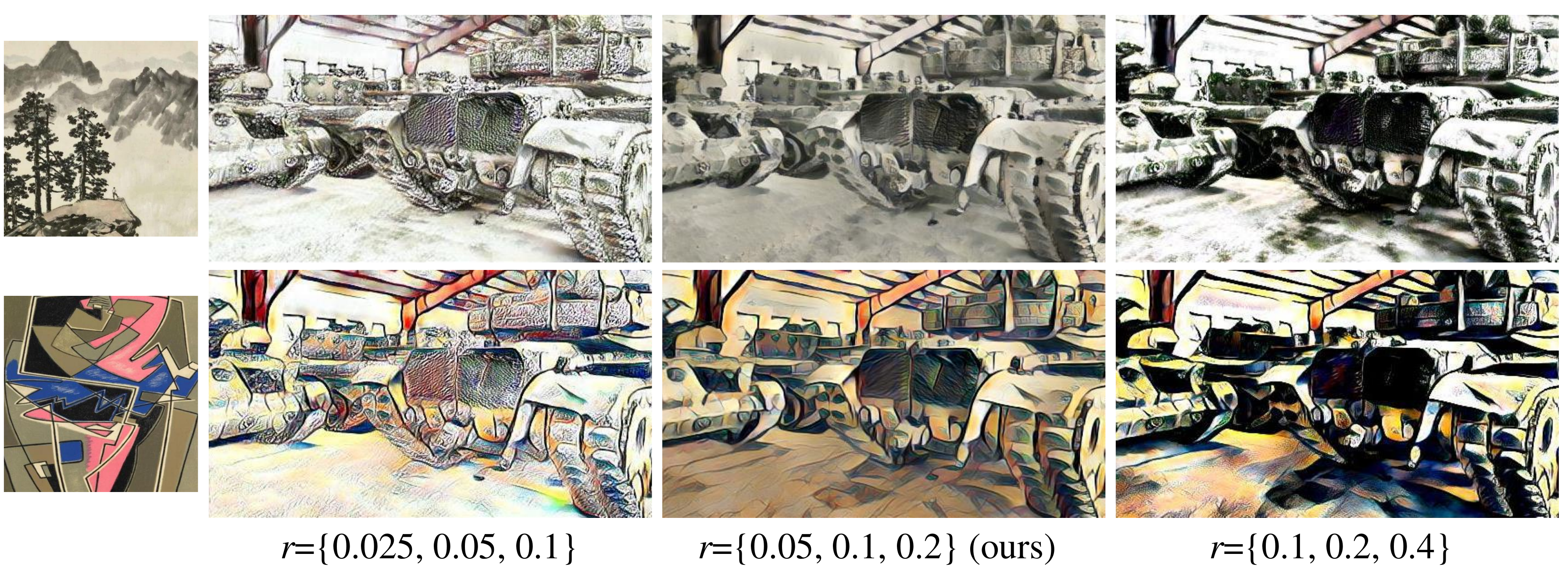}
\vspace{-1.5mm}
\caption{
\textbf{Ablation study on hyper-parameter $r$ in the point cloud aggregation.}
We compare the visual results of setting $r=\{0.025,0.05,0.1\}$, \textbf{$r=\{0.05,0.1,0.2\}$}, $r=\{0.1,0.2,0.4\}$.
We empirically determine to use $r=\{0.05,0.1,0.2\}$ for better visual quality.
}
\vspace{-3mm}
\label{fig:R}
\end{figure*}

\Paragraph{Radius $r$.}
\figref{R} shows the results of using different sets of radius parameters $r$ for our point cloud aggregation modules.
We empirically choose to use $r$=\{0.05,0.1,0.2\} for better visual quality.

\begin{figure*}[ht!]
\centering
\includegraphics[width=0.95\linewidth]{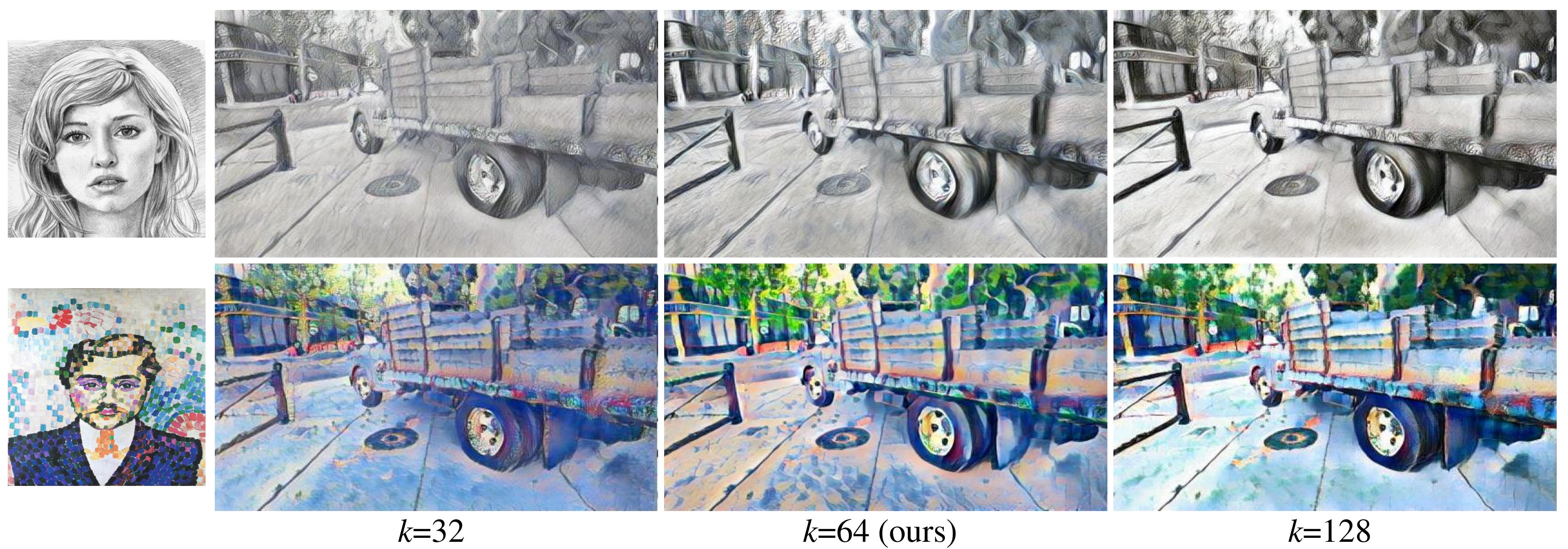}
\vspace{-1.5mm}
\caption{
\textbf{Ablation study on hyper-parameter $k$ in the point cloud aggregation.}
We compare the visual results of using $k=32/\mathbf{64}/128$, and empirically choose to use $k=64$ for better visual quality.
}
\vspace{-3mm}
\label{fig:K}
\end{figure*}

\Paragraph{Number of sampled points $k$.}
We conduct an ablation study to decide the parameter $k$.
\figref{K} shows the results of setting $k=32/64/128$.
We found that increasing the value of $k$ produces results with higher contrast.
We set $k$=64 since the results better match the style of the reference image.

\Paragraph{Quantitative analysis of applying point cloud aggregation modules.}
In \tabref{point2}, we provide the quantitative analysis to understand the impact of applying the point cloud aggregation modules on the consistency issue.
The results validate that using point aggregation modules improves both short-range and long-range consistency.

\begin{table}[ht!]
    \centering
    \footnotesize
	\caption{
	\textbf{Ablation study on point cloud aggregation.} We compute the short-range and long-range warping errors of the results generated by models with and without the point aggregation modules. 
	We validate that applying the point aggregation modules achieves better consistency across various novel views.
 }
 \vspace{1mm}

(a) Short-range consistency

	\vspace{1mm}
\begin{tabular}{l|ccccc} 
    \toprule
    Method & Truck & Playground & Train & M60 & Average \\
    \midrule
w/ aggregation	& 0.182	& 0.150	& 0.166	& 0.164	& \first{0.165} \\
w/o aggregation	& 0.187	& 0.159	& 0.167	& 0.164	& 0.168 \\
    \bottomrule
\end{tabular} 
	\vspace{1mm}
	
(b) Long-range consistency

\begin{tabular}{l|ccccc} 
    \toprule
    Method & Truck & Playground & Train & M60 & Average \\
    \midrule
w/ aggregation	& 0.590	& 0.332	& 0.409	& 0.434	& \first{0.428} \\
w/o aggregation	& 0.595	& 0.374	& 0.417	& 0.409	& 0.434 \\
    \bottomrule
\end{tabular} 

\label{tab:point2}
\end{table}

\subsubsection{PSNet for 3D Scene Stylization}
\begin{figure*}[ht!]
\centering
\includegraphics[width=0.95\linewidth]{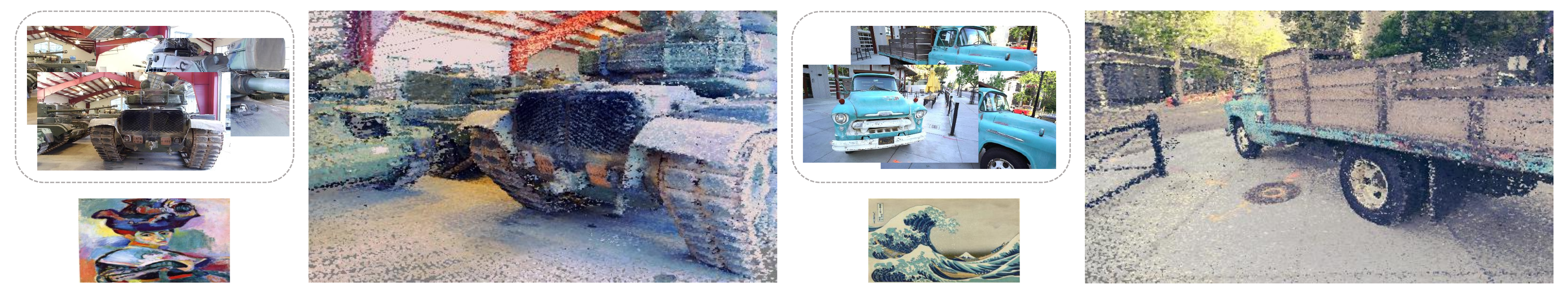}
\vspace{-1.5mm}
\caption{
\textbf{3D scene stylization results of PSNet.}
The PSNet~\cite{Cao_2020_WACV} generates projection artifacts and fails to produce desired stylization effect that matches the input reference image.
}
\vspace{-3mm}
\label{fig:psnet}
\end{figure*}

The PSNet \cite{Cao_2020_WACV} model aims to transfer the style of the point cloud. 
However, it is not applicable to our problem for two reasons.
First, PSNet requires per scene optimization on the ``RGB" point cloud. 
It fails to handle large-scale scenes in the real-world with more than 60M points, such as those in the Tanks and Temples dataset \cite{Knapitsch2017}. 
To make the PSNet framework applicable to our problem, we first use uniform sampling to reduce the number of RGB points in the point cloud to $1$M, then run PSNet framework to stylize the point cloud. 
We conduct the optimization process for the M60 and Truck scenes with 5000 iterations, which takes around 30 minutes for one specific combination of a scene and a reference image with desired style.
Compared to the runtime of the proposed method shown in \tabref{runtime}, the PSNet approach is time-consuming, thus limited for real-world applications.
After the construction of the RGB point cloud, we project the points to the 2D image plane to synthesize images at novel views.
As shown in \figref{psnet}, we observe that PSNet does not generate desired stylization effect that matches the input reference image. 
In addition, the PSNet produces projection artifacts that require post-processing schemes (\eg in-painting, smoothing) to refine the novel view synthesis results.

\subsubsection{Runtime Analysis}
In \tabref{runtime}, we show the training and inference time of the proposed method.
All the processes are conducted on a desktop machine equipped with a Nvidia Titan Xp GPU.
We note that after the point cloud transformation (3rd row) is completed, we can synthesize novel view images in near-real-time (\ie $17$ fps).
\begin{table}[ht!]\setlength{\tabcolsep}{5pt}
    \centering
    \caption{\textbf{Run-time analysis.} We present the training and inference time of each stage in the proposed method.}
    \label{tab:runtime}
    \begin{tabular}{l|c}
        \toprule
		Training time: decoder (seconds / per iteration) & $0.31$\\
		Training time: point cloud transformation module (seconds / per iteration) & $1.78$\\
		\midrule
		Inference time: constructing point cloud (seconds / per input image) &  $0.21$\\
        Inference time: stylizing point cloud (seconds / per scene) &  $0.74$\\
        Inference time: rendering novel view (seconds / per view)             & $0.06$\\
        \bottomrule
    \end{tabular}
\end{table}

\subsection{Implementation Details}

\begin{figure*}[ht!]
\centering
\includegraphics[width=0.5\linewidth]{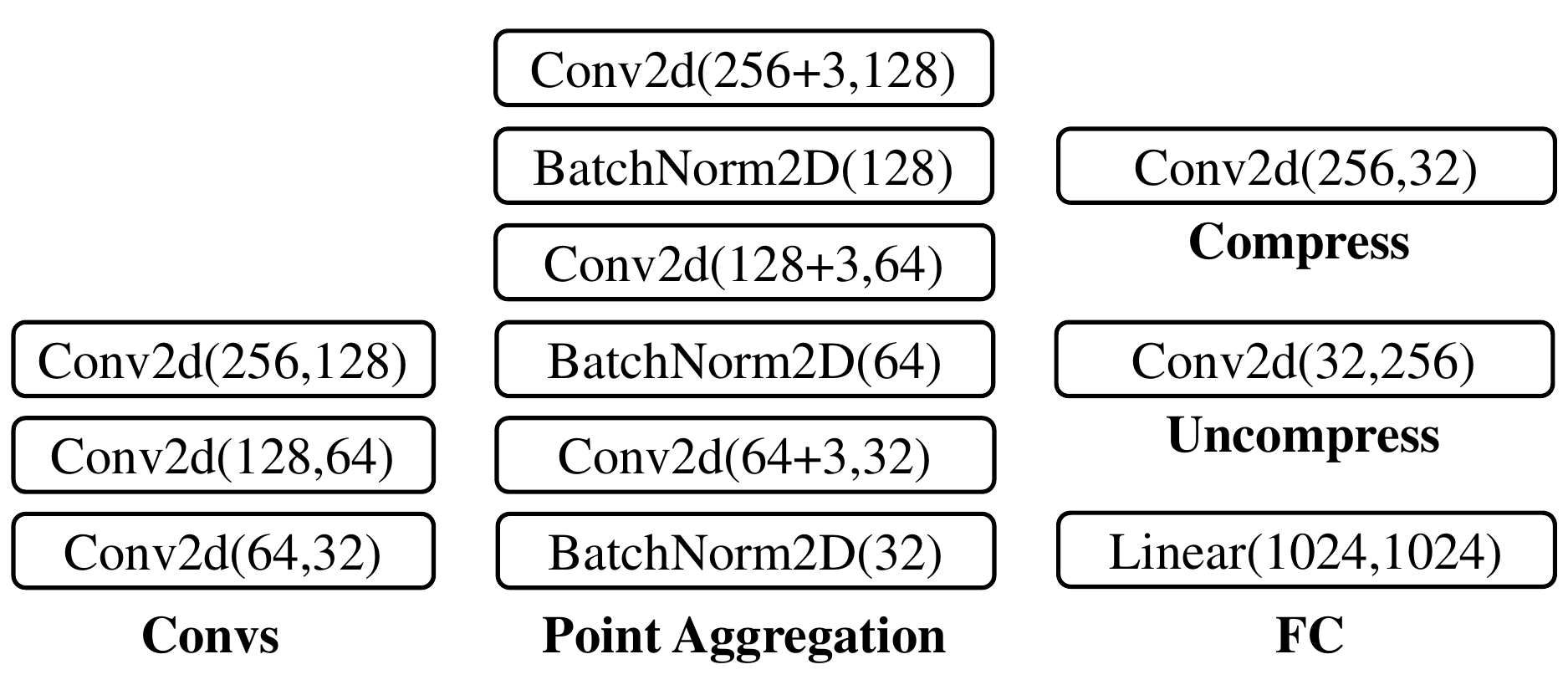}
\vspace{-1.5mm}
\caption{
\textbf{Network architecture.}
We present the network architecture of our point cloud transformation module illustrated in Figure 4 in the paper.
}
\vspace{-3mm}
\label{fig:arch}
\end{figure*}

\begin{figure*}[ht!]
\centering
\includegraphics[width=0.5\linewidth]{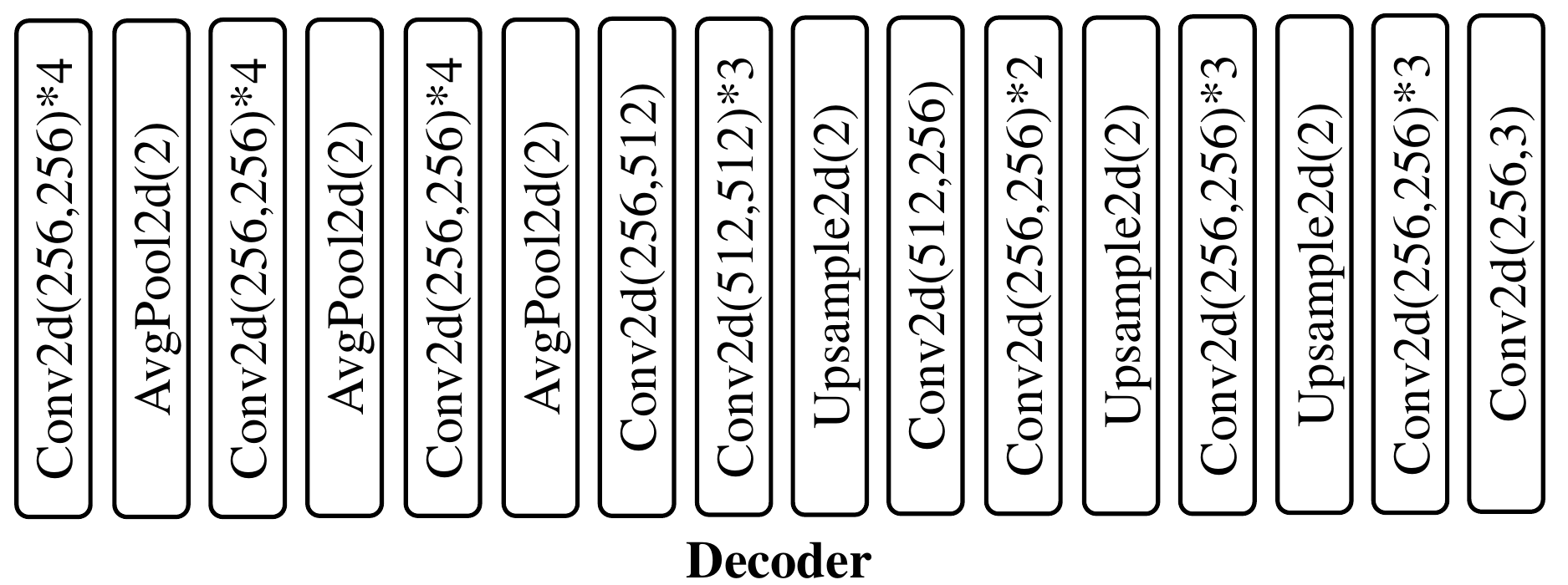}
\vspace{-1.5mm}
\caption{
\textbf{Decoder.}
We present the network architecture of our decoder module.
}
\vspace{-3mm}
\label{fig:dec}
\end{figure*}

\Paragraph{Network architecture.}
In \figref{arch}, we present the detailed architecture of each component in Figure 4 in the paper. We present the decoder architecture in \figref{dec}.

\Paragraph{Point cloud transformation.}
To reduce the computation cost in our point cloud transformation step, we employ the \emph{compress} and \emph{uncompress} operations in practice, described as follows.
The key is to reduce the feature dimension to accelerate the computation of the transformation matrix $\mathbf{T}$.
Specifically, we reduce (\ie compress) the feature dimension ($256\rightarrow32$) in the constructed point cloud $\{f^c_p\}^{P}_{p=1}$ through a MLP layer.
We then transform the constructed point cloud $\{f^c_p\}^{P}_{p=1}$ using the transformation matrix $\mathbf{T}$ of size $32\times32$.
Finally, we use a MLP layer to recover (\ie uncompress) the feature dimension ($32\rightarrow256$) to produce the transformed point cloud $\{f^d_p\}^{P}_{p=1}$ for the following novel view synthesis stage.
The process can be formulated as
\begin{equation}
    f^d_p = \mathrm{uncompress}(\mathbf{T}(\mathrm{compress}(f^c_p - \bar{f}^c))) + \bar{f}^s\hspace{5mm}\forall p \in [1,\cdots,P],
    \label{eq:transformation2}
\end{equation}
where $\bar{f}^c$ is the mean of the features in the point cloud $\{f^c_p\}^{P}_{p=1}$, and $\bar{f}^s$ is the mean of the style feature map $\mathbf{F}^s$.

\Paragraph{Point cloud aggregation.}
The number of points $P$ and feature dimension $c$ in each point cloud aggregation module is {\small$\{P, c\}: \{\approx2M, 256\} \rightarrow \{4096, 128\} \rightarrow \{2048, 64\} \rightarrow \{1024, 32\}$}.

\Paragraph{Novel view synthesis.}
Given a novel view $v$ with the camera pose $\{\mathbf{R}_v, t_v\}$ and intrinsic $\mathbf{K}_v$, we first project the features in our point cloud to the 2D image plane.
Specifically, we use the Pytorch3D \cite{ravi2020pytorch3d} point cloud renderer for the projection of features.
We set the size of the z-buffer as 128 and the points are splatted to a region with radius of 2 pixels.
We then use a decoder presented in \figref{dec} to synthesize the final image from the projected 2D feature map.

\Paragraph{Training.}
We implement our system in PyTorch, and use the Adam optimizer~\cite{adam} with $\beta_1 = 0.9$, $\beta_2 = 0.9999$ for all network training.
We first train the decoder module for $50$K iterations with a batch size of $1$ and learning rate of $0.0001$.
Following the WCT approach~\cite{WCT-NIPS-2017}, the $\ell1$ reconstruction loss illustrated in Line 417 in the paper is the combination of the pixel reconstruction loss and feature loss.
Particularly, the feature loss is computed using the features of a pre-trained VGG-19 network, including \{conv1\_2, conv2\_2, conv3\_2, conv4\_2, conv5\_2\}.
We then train the transformation module for $50$K iterations with a batch size of $1$ and learning rate of $0.0001$.
The content loss described in Eq. (2) in the paper is computed by the features of layer relu4\_1, while the style loss is computed by \{relu1\_1, relu2\_1, relu3\_1, relu4\_1\}.
The weight $\lambda$ for the style loss is set to $0.02$.
To improve the training efficiency, we uniformly down-sample the constructed point cloud to $600$K features for each scene, and use all the features in the point cloud during the testing time.

\subsection{Explicit vs. Implicit Representations}
While implicit representation-based approaches~\cite{barron2021mipnerf,boss2020nerd,chanmonteiro2020piGAN,chen2021mvsnerf,kangle2021dsnerf,devries2021unconstrained,du2020nerflow,Gafni_2021_CVPR,Gaoportraitnerf,garbin2021fastnerf,guo2020osf,kosiorek2021nerfvae,li2021neural3dvideo,li2020neural,autoint2021,liu2020neural,liu2021editing,Lombardi2021MixtureOV,martinbrualla2020nerfw,meng2021gnerf,nerf,neff2021donerf,niemeyer2021campari,Niemeyer2020GIRAFFE,noguchi2021neural,ost2020neuralscenegraphs,park2020nerfies,park2021hypernerf,peng2021animatable,peng2021neural,pumarola2020d,raj2021pva,Rebain_2021_CVPR,reiser2021kilonerf,rematasICML21,Schwarz2020NEURIPS,nerv2021,su2021anerf,Sucararxiv2021,tancik2020meta,Tretschk20arxiv_NR,grf2020,wang2021ibrnet,wang2020learning,wang2021nerfmm,Wizadwongsa2021NeX,Xian2021,xie2021fignerf,yen2020inerf,yu2021plenoctrees,yu2020pixelnerf,kaizhang2020,nerfactor,ZhiICCV2021} produce high-quality (non-stylized) novel view synthesis results, we choose to leverage explicit representations due to the practical considerations that support real-world VR/AR applications: \emph{efficiency} and \emph{scalability}.
Specifically, the NeRF++ method~\cite{kaizhang2020} is designed for complex unbounded 3D scenes.
Nevertheless, it takes $24$ hours to reconstruct a \emph{particular} scene, and $30$ seconds to render a 546$\times$980 image.
Moreover, the NeRF++-based framework produces blurry stylization results due to the inconsistency issue, as shown in~\figref{nerf}.
Although there are recent efforts~\cite{garbin2021fastnerf,autoint2021,liu2020neural,Lombardi2021MixtureOV,neff2021donerf,Rebain_2021_CVPR,reiser2021kilonerf,yu2021plenoctrees} to accelerate the rendering process, these schemes are limited to single 3D objects or bounded 3D scenes.
In contrast, the proposed method is efficient, and renders the stylized novel views in near-real-time, as presented in~\tabref{runtime}.
Furthermore, the proposed method is more scalable than the NeRF++-based approaches since it handles arbitrary unbounded scenes and styles with a single trained model.

\subsection{Limitations and Future Direction}

\begin{figure*}[ht!]
\centering
\includegraphics[width=0.4\linewidth]{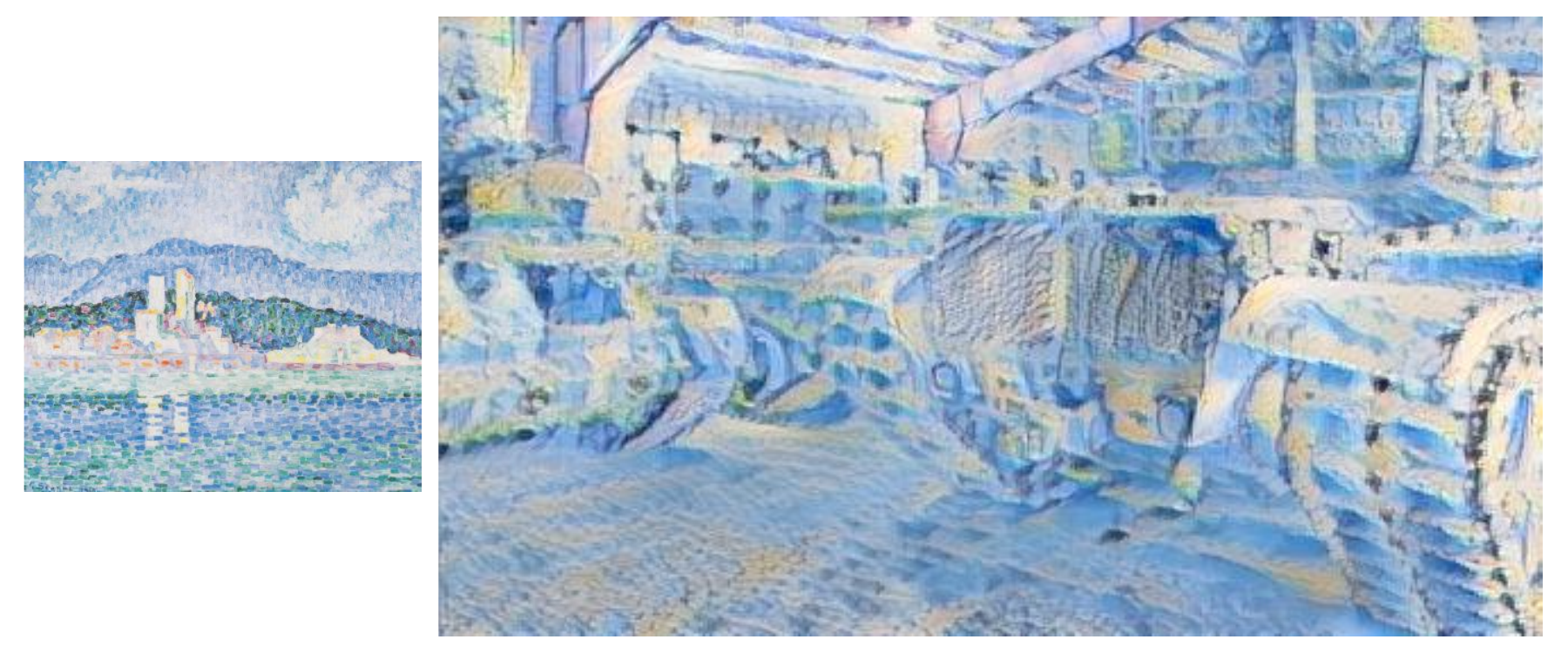}
\vspace{-1.5mm}
\caption{
\textbf{Limitations.}
Our model is not aware of individual objects in the scene during the stylization process, thus fail to transfer the style of a particular part of the reference image to the specific object/region in the scene.
}
\vspace{-3mm}
\label{fig:failure}
\end{figure*}

We discuss the limitation of our method, which we plan to explore in the future work as follows. 
First, as shown in \figref{failure}, our 3D scene stylization approach is not aware of the objects in the scene.
As a result, we cannot transfer the style of the particular part of the style image to the specific object/region of the 3D scene.
Second, the proposed approach cannot significantly modify the geometry of the scene during the stylization process since 1) our point cloud is built according to the 3D proxy of the original scene and 2) we only transform the features in our point cloud, but not adjust the location of each point.
In the future, we plan to explore the solution that is 1) 3D object-aware and 2) capable of modulating the geometry of the 3D scene to match the desired style.

\end{document}